\documentclass[10pt,twocolumn,letterpaper]{article}

\usepackage{cvpr}
\usepackage{times}
\usepackage{epsfig}
\usepackage{graphicx}
\usepackage{amsmath}
\usepackage{amssymb}
\usepackage{subfiles}
\usepackage{color}
\usepackage{enumitem}
\usepackage{dsfont}
\usepackage{booktabs}
\usepackage{float}
\usepackage{capt-of,etoolbox}
\usepackage{graphicx}
\usepackage{caption}
\usepackage{subcaption}
\usepackage{multirow}
\usepackage{arydshln}

\definecolor{MyDarkBlue}{rgb}{0.08,0.02,0.5}
\definecolor{MyLightBlue}{rgb}{0,0.08,0.8}
\definecolor{MyDarkGreen}{rgb}{0.02,0.50,0.02}
\definecolor{MyDarkRed}{rgb}{0.7,0.02,0.02}
\definecolor{MyDarkOrange}{rgb}{0.40,0.2,0.02}
\definecolor{MyDarkMagenta}{rgb}{0.337,0,0.827}
\definecolor{MyDarkGray}{rgb}{0.5,0.5,0.5}
\newcommand{\gray}[1]{\textcolor{MyDarkGray}{#1}}
\newcommand{\tbfit}[1]{\textbf{\textit{#1}}}
\newcommand{\tbfu}[1]{\textbf{\underline{#1}}}

\renewcommand{\paragraph}[1]{\vspace{5px} \noindent \textbf{#1} \ \ }

\usepackage[pagebackref=true,breaklinks=true,letterpaper=true,colorlinks,bookmarks=false,draft]{hyperref}

\cvprfinalcopy 

\def\cvprPaperID{299} 

\begin{document}

\title{The Unreasonable Effectiveness of Deep Features as a Perceptual Metric}

\author{Richard Zhang$^{1}$ \hspace{3mm} Phillip Isola$^{12}$ \hspace{3mm} Alexei A. Efros$^{1}$\\
$^{1}$UC Berkeley \hspace{3mm} $^{2}$OpenAI\\
{\tt\small \{rich.zhang, isola, efros\}@eecs.berkeley.edu}
\and
Eli Shechtman$^{3}$ \hspace{3mm} Oliver Wang$^{3}$\\
$^{3}$Adobe Research\\
{\tt\small \{elishe,owang\}@adobe.com}
}

\twocolumn[{%
\vspace{-45px}
\renewcommand\twocolumn[1][]{#1}%
\maketitle
\centering
\includegraphics[width=1.\linewidth]{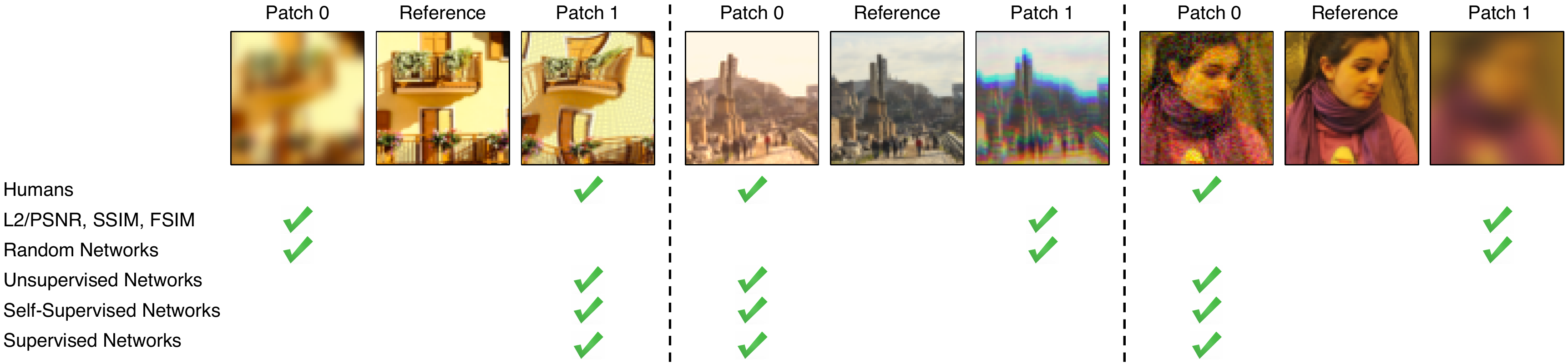}
\vspace{-16px}
\captionof{figure}{\label{fig:fig1}
\textbf{Which patch (left or right) is ``closer" to the middle patch in these examples?} In each case, the traditional metrics (L2/PSNR, SSIM, FSIM) disagree with human judgments. 
But deep networks, even across architectures (Squeezenet~\cite{iandola2016squeezenet}, AlexNet~\cite{krizhevsky2014one}, VGG~\cite{simonyan2014very}) and supervision type (supervised~\cite{russakovsky2015imagenet}, self-supervised~\cite{donahue2016adversarial, noroozi2016unsupervised,pathak2017learning,zhang2017split}, and even unsupervised~\cite{krahenbuhl2015data}), provide an \textit{emergent embedding} which agrees surprisingly well with humans. We further calibrate existing deep embeddings on a large-scale database of perceptual judgments; models and data can be found at \url{https://www.github.com/richzhang/PerceptualSimilarity}.}
\vspace{10px}
}]



\begin{abstract}
\vspace{-8px}

While it is nearly effortless for humans to quickly assess the perceptual similarity between two images, the  underlying processes are thought to be quite complex. Despite this, the most widely used perceptual metrics today, such as PSNR and SSIM, are simple, shallow functions, and fail to account for many nuances of human perception. Recently, the deep learning community has found that features of the VGG network trained on ImageNet classification has been remarkably useful as a training loss for image synthesis. But how perceptual are these so-called ``perceptual losses"? What elements are critical for their success? To answer these questions, we introduce a new dataset of human perceptual similarity judgments. We systematically evaluate deep features across different architectures and tasks and compare them with classic metrics. We find that deep features outperform all previous metrics by large margins on our dataset. More surprisingly, this result is not restricted to ImageNet-trained VGG features, but holds across different deep architectures and levels of supervision (supervised, self-supervised, or even unsupervised). Our results suggest that perceptual similarity is an emergent property shared across deep visual representations.

\end{abstract}

\vspace{-10px}


\begin{figure*}
\centering
\begin{subfigure}{.485\textwidth}
  \centering
  \includegraphics[width=1.\linewidth]{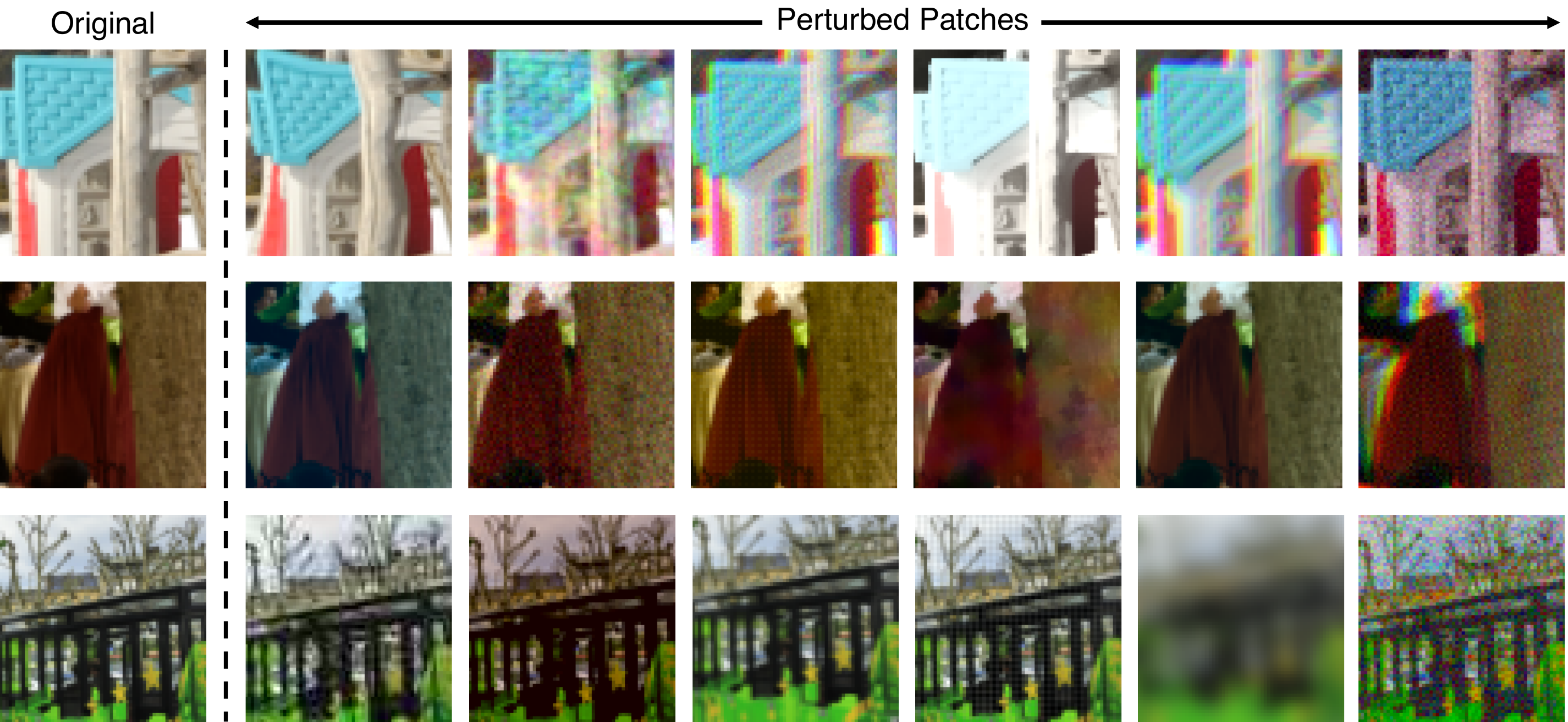}
  \caption{\textbf{Traditional}}
  \label{fig:pert_low}
\end{subfigure}
\hspace{8px}
\begin{subfigure}{.485\textwidth}
  \centering
  \includegraphics[width=1.\linewidth]{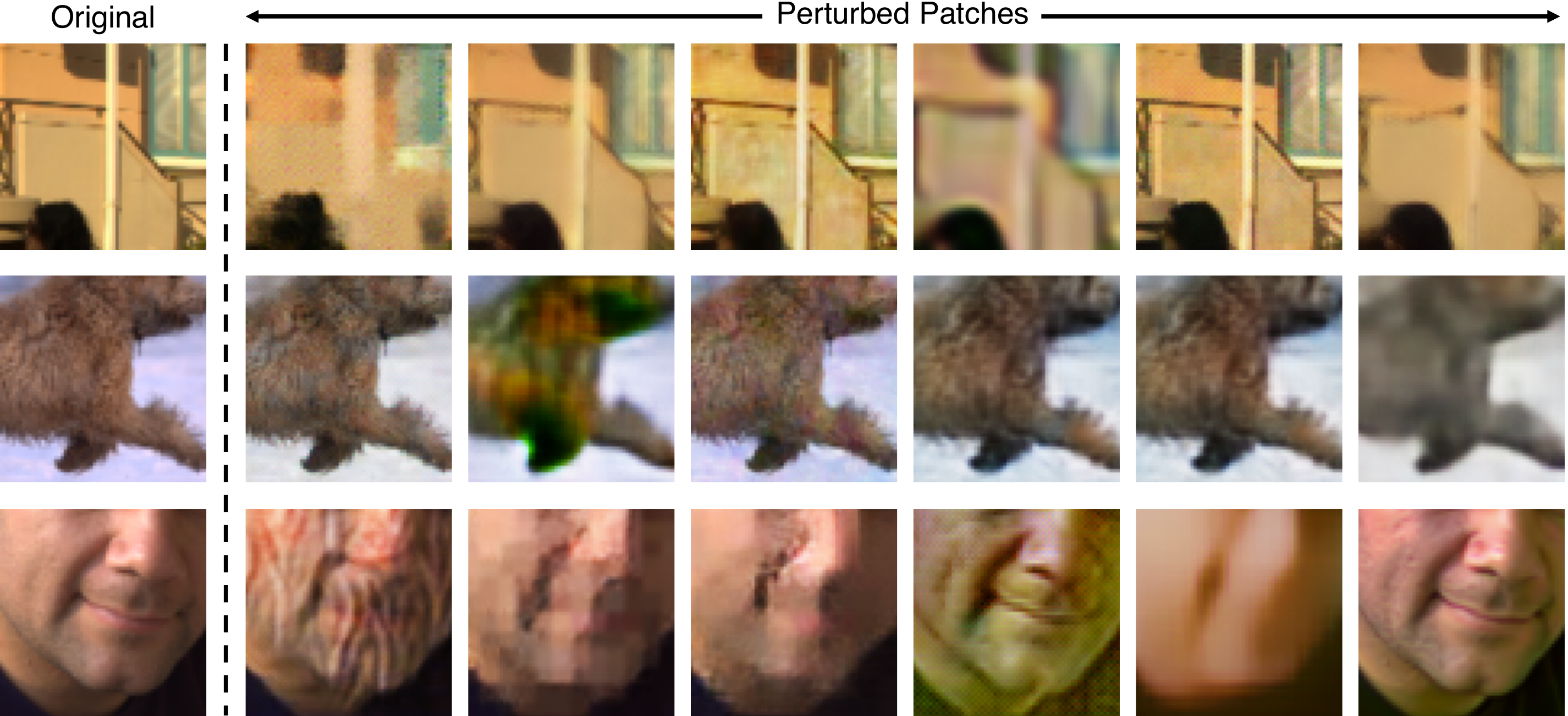}
  \caption{\textbf{CNN-based}}
  \label{fig:pert_high}
\end{subfigure}
\vspace{-2mm}
\caption{\textbf{Example distortions.} We show example distortions using our (a) traditional and (b) CNN-based methods.}
\vspace{-4mm}
\label{fig:pert}
\end{figure*}

\section{Motivation}

The ability to compare data items is perhaps the most fundamental operation underlying all of computing.  In many areas of computer science it does not pose much difficulty: one can use Hamming distance to compare binary patterns, edit distance to compare text files, Euclidean distance to compare vectors, etc.  The unique challenge of computer vision is that even this seemingly simple task of comparing visual patterns remains a wide-open problem.  Not only are visual patterns very high-dimensional and highly correlated, but, the very notion of visual similarity is often subjective, aiming to mimic human visual perception.  For instance, in image compression, the goal is for the compressed image to be indistinguishable from the original by a human observer, irrespective of the fact that their pixel representations might be very different. 

Classic per-pixel measures, such as $\ell_2$ Euclidean distance, commonly used for regression problems, or the related Peak Signal-to-Noise Ratio (PSNR), are insufficient for assessing structured outputs such as images, as they assume pixel-wise independence. A well-known example is that blurring causes large perceptual but small $\ell_2$ change.

What we would really like is a ``perceptual distance," which measures how similar are two images in a way that coincides with human judgment. This problem has been a longstanding goal,
and there have been numerous perceptually motivated distance metrics proposed, such as SSIM~\cite{wang2004image}, MSSIM~\cite{wang2003multiscale}, FSIM~\cite{zhang2011fsim}, and HDR-VDP~\cite{mantiuk2011hdr}.

However, constructing a perceptual metric is challenging, because human judgments of similarity (1) depend on high-order image structure \cite{wang2004image}, (2) are context-dependent \cite{Goodman1972,medin1993respects,markman2005nonintentional}, and (3) may not actually constitute a distance metric \cite{tversky1977features}.
The crux of (2) is that there are many different ``senses of similarity" that we can simultaneously hold in mind: is a red circle more similar to a red square or to a blue circle? Directly fitting a function to human judgments may be intractable due the the context-dependent and pairwise nature of the judgments (which compare the similarity between \textit{two} images). Indeed, we show in this paper a negative result where this approach fails to generalize, even when trained on a large-scale dataset containing many distortion types.

Instead, might there be a way to learn a notion of perceptual similarity without directly training for it? The computer vision community has discovered that internal activations of deep convolutional networks, though trained on a high-level image classification task, are often surprisingly useful as a representational space for a much wider variety of tasks. For example, features from the VGG architecture~\cite{simonyan2014very} have been used on tasks such as neural style transfer~\cite{gatys2016image}, image superresolution~\cite{johnson2016perceptual}, and conditional image synthesis~\cite{dosovitskiy2016generating,chen2017photographic}. 
These methods measure distance in VGG feature space as a ``perceptual loss" for image regression problems~\cite{johnson2016perceptual,dosovitskiy2016generating}.

But how well do these ``perceptual losses'' actually correspond to human visual perception? How do they compare to traditional perceptual image evaluation metrics? Does the network architecture matter? Does it have to be trained on the ImageNet classification task, or would other tasks work just as well?  Do the networks need to be trained at all?

\begin{table*}[t!]
\begin{center}
\scalebox{0.83} {
\begin{tabular}{ c c c c c c c c c }
\textbf{Dataset} & \textbf{$\#$ Input Imgs/} & \textbf{Input} & \textbf{Num} & \textbf{Distort.} & \textbf{$\#$} & \textbf{$\#$ Distort.} & \textbf{$\#$ Judg-} & \textbf{Judgment} \\ 
& \textbf{Patches} & \textbf{Type} & \textbf{Distort.} & \textbf{Types} & \textbf{Levels} & \textbf{Imgs/Patches} & \textbf{ments} & \textbf{Type} \\
\toprule
LIVE~\cite{sheikh2006statistical} & 29 & images & 5 & traditional & continuous & .8k & 25k & MOS \\ 
CSIQ~\cite{larson2010most} & 30 & images & 6 & traditional & 5 & .8k & 25k & MOS \\ 
TID2008~\cite{ponomarenko2009tid2008} & 25 & images & 17 & traditional & 4 & 2.7k & 250k & MOS \\
TID2013~\cite{ponomarenko2015image} & 25 & images & 24 & traditional & 5 & 3.0k & 500k & MOS \\ 
\midrule
BAPPS (2AFC--Distort) & 160.8k & $64\times 64$ patch & 425 & trad + CNN & continuous & 321.6k & 349.8k & 2AFC \\
BAPPS (2AFC--Real alg) & 26.9k & $64\times 64$ patch & -- & alg outputs & -- & 53.8k & 134.5k & 2AFC  \\ \hdashline 
BAPPS (JND--Distort) & 9.6k & $64\times 64$ patch & 425 & trad. + CNN & continuous & 9.6k & 28.8k & Same/Not same  \\ 

\bottomrule
\end{tabular}
}
\vspace{5px}
\vspace{-2mm}
\caption{\textbf{Dataset comparison.} A primary differentiator between our proposed Berkeley-Adobe Perceptual Patch Similarity (BAPPS) dataset and previous work is scale of distortion types. We provide human perceptual judgments on distortion set using uncompressed images from~\cite{fivek,dang2015raise}. Previous datasets have used a small number of distortions at discrete levels. We use a large number of distortions (created by sequentially composing atomic distortions together) and sample continuously. For each input patch, we corrupt it using two distortions and ask for a few human judgments (2 for train, 5 for test set) per pair. This enables us to obtain judgments on a large number of patches. Previous databases summarize their judgments into a mean opinion score (MOS); we simply report pairwise judgments (two alternative force choice). In addition, we provide judgments on outputs from \textit{real algorithms}, as well as a same/not same Just Noticeable Difference (JND) perceptual test.
\vspace{-8mm}
}
\label{tab:dataset_comp}
\end{center}
\end{table*}

In this paper, we evaluate these questions on a new large-scale database of human judgments, and arrive at several surprising conclusions. We find that internal activations of networks trained for high-level classification tasks, even across network architectures~\cite{iandola2016squeezenet,krizhevsky2012imagenet,simonyan2014very} and no further calibration, do indeed correspond to human perceptual judgments. In fact, they correspond far better than the commonly used metrics like SSIM and FSIM~\cite{wang2004image, zhang2011fsim}, which were not designed to handle situations where spatial ambiguities are a factor~\cite{sampat2009complex}. Furthermore, the best performing self-supervised networks, including BiGANs~\cite{donahue2016adversarial}, cross-channel prediction~\cite{zhang2017split}, and puzzle solving~\cite{noroozi2016unsupervised} perform just as well at this task, even without the benefit of human-labeled training data. Even a simple unsupervised network initialization with stacked k-means~\cite{krahenbuhl2015data} beats the classic metrics by a large margin! 
This illustrates an \textit{emergent property} shared across networks, even across architectures and training signals. Importantly, however, having \textit{some} training signal appears crucial -- a randomly initialized network achieves much lower performance.

Our study is based on a newly collected perceptual similarity dataset, using a large set of distortions and real algorithm outputs. It contains both traditional distortions, such as contrast and saturation adjustments, noise patterns, filtering, and spatial warping operations, and CNN-based algorithm outputs, such as autoencoding, denoising, and colorization, produced by a variety of architectures and losses. Our dataset is richer and more varied than previous datasets of this kind~\cite{ponomarenko2015image}. We also collect judgments on outputs from real algorithms for the tasks of superresolution, frame interpolation, and image deblurring, which is especially important as these are the real-world use cases for a perceptual metric.
We show that our data can be used to ``calibrate" existing networks, by learning a simple linear scaling of layer activations, to better match low-level human judgments.

Our results are consistent with the hypothesis that perceptual similarity is not a special function all of its own, but rather a {\em consequence} of visual representations tuned to be predictive about important structure in the world. Representations that are effective at semantic prediction tasks are also representations in which Euclidean distance is highly predictive of perceptual similarity judgments.

Our contributions are as follows:

\begin{itemize}[noitemsep,nolistsep]
    \item We introduce a large-scale, highly varied, perceptual similarity dataset, containing 484k human judgments. Our dataset not only includes parameterized distortions, but also real algorithm outputs. We also collect judgments on a different perceptual test, just noticeable differences (JND).
    
    \item We show that deep features, trained on supervised, self-supervised, and unsupervised objectives alike, model low-level perceptual similarity surprisingly well, outperforming previous, widely-used metrics.
    
    \item We demonstrate that network architecture alone does not account for the performance: untrained nets achieve much lower performance.
    
    \item With our data, we can improve performance by ``calibrating" feature responses from a pre-trained network.
    
\end{itemize}

\paragraph{Prior work on datasets}
%
In order to evaluate existing similarity measures, a number of datasets have been proposed. Some of the most popular are the LIVE~\cite{sheikh2006statistical}, TID2008~\cite{ponomarenko2009tid2008},  CSIQ~\cite{larson2010most}, and TID2013~\cite{ponomarenko2015image} datasets. These datasets are referred to Full-Reference Image Quality Assessment (FR-IQA) datasets and have served as the de-facto baselines for development and evaluation of similarity metrics. A related line of work is on No-Reference Image Quality Assessment (NR-IQA), such as AVA~\cite{murray2012ava} and LIVE In the Wild~\cite{ghadiyaram2016massive}. These datasets investigate the ``quality" of individual images by themselves, without a reference image.
We collect a new dataset that is complementary to these: it contains a substantially larger number of distortions, including some from newer, deep network based outputs, as well as geometric distortions. Our dataset is focused on perceptual similarity, rather than quality assessment. Additionally, it is collected on patches as opposed to full images, in the wild, with a different experimental design (more details in Sec~\ref{sec:methods}).

\paragraph{Prior work on deep networks and human judgments} Recently, advances in DNNs have motivated investigation of applications in the context of visual similarity and image quality assessment. 
Kim and Lee~\cite{kim2017deep} use a CNN to predict visual similarity by training on low-level differences. Concurrent work by Talebi and Milanfar~\cite{talebi2017learned,talebi2018nima} train a deep network in the context of NR-IQA for image aesthetics. Gao et al.~\cite{gao2017deepsim} and Amirshahi et al.~\cite{ali2017image} propose techniques involving leveraging internal activations of deep networks (VGG and AlexNet, respectively) along with additional multiscale post-processing. In this work, we conduct a more in-depth study across different architectures, training signals, on a new, large scale, highly-varied dataset.

Recently, Berardino et al.~\cite{berardino2017eigen} train networks on perceptual similarity, and importantly, assess the ability of deep networks to make predictions on a \textit{separate} task -- predicting most and least perceptually-noticeable directions of distortion. Similarly, we not only assess image patch similarity on parameterized distortions, but also test generalization to real algorithms, as well as generalization to a separate perceptual task -- just noticeable differences.

\section{Berkeley-Adobe Perceptual Patch Similarity (BAPPS) Dataset}
\label{sec:methods}

To evaluate the performance of different perceptual metrics, we collect a large-scale highly diverse dataset of perceptual judgments using two approaches.
Our main data collection employs a two alternative forced choice (2AFC) test, that asks which of two distortions is more similar to a reference.
This is validated by a second experiment where we perform a just noticeable difference (JND) test, which asks whether two patches -- one reference and one distorted -- are the same or different. These judgments are collected over a wide space of distortions and real algorithm outputs.

\begin{table*}
\centering
\parbox[t][][t]{.48\linewidth}{
\resizebox{\linewidth}{!}{
\begin{tabular}{ c c }
\textbf{Sub-type} & \textbf{Distortion type} \\ \hline
Photometric & lightness shift, color shift, contrast, saturation \\ \hline
 & uniform white noise, Gaussian white, pink, \\
Noise & \& blue noise, Gaussian colored (between \\ 
 & violet and brown) noise, checkerboard artifact \\ \hline
 Blur & Gaussian, bilateral filtering \\ \hline
Spatial & shifting, affine warp, homography, \\
& linear warping, cubic warping, ghosting, \\
& chromatic aberration, \\ \hline
Compression & jpeg \\ \hline
\end{tabular} } } \hfill
\parbox[t][][t]{.48\linewidth}{
\resizebox{\linewidth}{!}{
\begin{tabular}{ c c }
\textbf{Parameter type} & \textbf{Parameters} \\ \hline
Input & null, pink noise, white noise, \\
corruption & color removal, downsampling \\ \hline
 & \# layers, \# skip connections, \\
Generator & \# layers with dropout, force skip connection \\
network & at highest layer, upsampling method, \\
architecture & normalization method, first layer stride \\ 
& \# channels in $1^{st}$ layer, max \# channels \\ \hline
Discriminator & number of layers \\ \hline
Loss/Learning & weighting on oixel-wise ($\ell_1$), VGG, \\
& discriminator losses, learning rate \\ \hline
\end{tabular} } }
\vspace{-2mm}
\caption{\label{tab:distortions}
\textbf{Our distortions.} Our traditional distortions (left) are performed by basic low-level image editing operations. We also sequentially compose them to better explore the space. Our CNN-based distortions (right) are formed by randomly varying parameters such as task, network architecture, and learning parameters. The goal of the distortions is to mimic plausible distortions seen in real algorithm outputs.}
\vspace{-4mm}
\end{table*}

\subsection{Distortions}

\paragraph{Traditional distortions} We create a set of ``traditional" distortions consisting of common operations performed on the input patches, listed in Table~\ref{tab:distortions} (left). In general, we use photometric distortions, random noise, blurring, spatial shifts and corruptions, and compression artifacts. We show qualitative examples of our traditional distortions in Figure \ref{fig:pert}. The severity of each perturbation is parameterized - for example, for Gaussian blur, the kernel width determines the amount of corruption applied to the input image. We also compose pairs of distortions sequentially to increase the overall space of possible distortions. In total, we have 20 distortions and 308 sequentially composed distortions.

\paragraph{CNN-based distortions} To more closely simulate the space of artifacts that can arise from deep-learning based methods, we create a set of distortions created by neural networks. We simulate possible algorithm outputs by exploring a variety of tasks, architectures, and losses, as shown in Table \ref{tab:distortions} (right). Such tasks include autoencoding, denoising, colorization, and superresolution. All of these tasks can be achieved by applying the appropriate corruption to the input. In total, we generated 96 ``denoising autoencoders" and use these as CNN-based distortion functions. We train each of these  networks on the 1.3M ImageNet dataset~\cite{russakovsky2015imagenet} for 1 epoch. The goal of each network is not to solve the task per se, but rather to explore common artifacts that plague the outputs of deep learning based methods.

\paragraph{Distorted image patches from real algorithms}
The true test of an image assessment algorithm is on real problems and real algorithms. 
We gather perceptual judgments using such outputs. Data on real algorithms is more limited, as each application will have their own unique properties. For example, different colorization methods will not show much structural variation, but will be prone to effects such as color bleeding and color variation. On the other hand, superresolution will not have color ambiguity, but may see larger structural changes from algorithm to algorithm.

\paragraph{Superresolution} We evaluate results from the 2017 NTIRE workshop~\cite{Agustsson_2017_CVPR_Workshops}. We use 3 tracks from the workshop -- $\times2$, $\times3$, $\times4$ upsampling rates using ``unknown" downsampling to create the input images. Each track had approximately 20 algorithm submissions. We also evaluate several additional methods, including bicubic upsampling, and four of the top performing deep superresolution methods~\cite{kim2016accurate,wang2015deep,ledig2016photo,sajjadi2016enhancenet}. A common qualitative way of presenting superresolution results is zooming into specific patches and comparing differences. As such, we sample random $64\times 64$ triplets from random locations of images in the Div2K~\cite{Agustsson_2017_CVPR_Workshops} dataset -- the ground truth high-resolution image, along with two algorithm outputs.

\paragraph{Frame interpolation} We sample patches from different frame interpolation algorithms, including three variants of flow-based interpolation~\cite{liu2009beyond}, CNN-based interpolation~\cite{Niklaus_ICCV_2017}, and phase-based interpolation~\cite{meyer2015phase} on the Davis Middleburry dataset~\cite{scharstein2002taxonomy}. Because artifacts arising from frame interpolation may occur at different scales, we randomly rescale the image before sampling a patch triplet.

\paragraph{Video deblurring} We sample from the video deblurring dataset~\cite{Su_2017_CVPR}, along with deblurring outputs from Photoshop Shake Reduction, Weighted Fourier Aggregation~\cite{delbracio2015hand}, and three variants of a deep video deblurring method~\cite{Su_2017_CVPR}.

\paragraph{Colorization} We sample patches using random scales on the colorization task, on images from the ImageNet dataset~\cite{russakovsky2015imagenet}. The algorithms are from pix2pix~\cite{isola2015learning}, Larsson et al.~\cite{larsson2016learning}, and variants from Zhang et al.~\cite{zhang2016colorful}.

\subsection{Psychophysical Similarity Measurements}

\paragraph{2AFC similarity judgments}
We randomly select an image patch $x$ and apply two distortions to produce patches $x_0, x_1$. We then ask a human which is closer to the original patch $x$, and record response $h \in \{0,1\}$. On average, people spent approximately 3 seconds per judgment. Let $\mathcal{T}$ denote our dataset of patch triplets $(x, x_0, x_1, h)$.

A comparison between our dataset and previous datasets is shown in Table~\ref{tab:dataset_comp}. 
Previous datasets have focused on collecting large numbers of human judgments for a few images and distortion types. 
For example, the largest dataset, TID2013~\cite{ponomarenko2015image}, has 500k judgments on 3000 distortions (from 25 input images with 24 distortions types, each sampled at 5 levels). 
We provide a complementary dataset that focuses instead on a large number of distortions types. In, addition, we collect judgments on a large number of $64\times 64$ patches rather than a small number of images.
There are three reasons for this. 
First, the space of full images is extremely large, which makes it much harder to cover a reasonable portion of the domain with judgments (even $64\times 64$ color patches represent an intractable 12k-dimensional space).
Second, by choosing a smaller patch size, we focus on lower-level aspects of similarity, to mitigate the effect of differing ``respects of similarity" that may be influenced by high-level semantics \cite{medin1993respects}. Finally, modern methods for image synthesis train deep networks with patch-based losses (implemented as convolutions)~\cite{chen2017photographic,isola2017image}. 
Our dataset consists of over 161k patches, derived from the MIT-Adobe 5k dataset~\cite{fivek} (5000 uncompressed images) for training, and the RAISE1k dataset~\cite{dang2015raise} for validation.

To enable large-scale collection, our data is collected ``in-the-wild" on Amazon Mechanical Turk, as opposed to a controlled lab setting. 
Crump et al.~\cite{crump2013evaluating} show that AMT can be reliably used to replicate many psychophysics studies, despite the inability to control all environmental factors.
We ask for 2 judgments per example in our ``train" set and 5 judgments in our ``val" sets.
Asking for fewer judgments enables us to explore a larger set of image patches and distortions. We add sentinels which consist of pairs of patches with obvious deformations, e.g., a large amount of Gaussian noise vs a small amount of Gaussian noise. Approximately $~90\%$ of Turkers were able to correctly pass at least $93\%$ of the sentinels (14 of 15), indicating that they understood the task and were paying attention.
We choose to use a larger number of distortions than prior datasets.

\begin{table}[t]
\begin{center}
\scalebox{0.85} {
\begin{tabular}{c c c c c }
\multirow{2}[2]{*}{\textbf{Dataset}} & \textbf{Data} & \textbf{Train/} & \textbf{$\#$ Ex-} & \textbf{$\#$ Judge} \\
& \textbf{source} & \textbf{Val} & \textbf{amples} & \textbf{/Example} \\ \midrule \midrule
Traditional & ~\cite{fivek} & Train & 56.6k & 2 \\
CNN-based & ~\cite{fivek} & Train & 38.1k & 2 \\
Mixed & ~\cite{fivek} & Train & 56.6k & 2 \\ \midrule
\textbf{2AFC--Distort [Trn]} & -- & Train & 151.4k & 2 \\ \midrule \midrule
Traditional & ~\cite{dang2015raise} & Train & 4.7k & 5 \\
CNN-based & ~\cite{dang2015raise} & Train & 4.7k & 5 \\ \midrule 
\textbf{2AFC--Distort [Val]} & -- & Val & 9.4k & 5\\ \midrule \midrule
Superres & ~\cite{Lim_2017_CVPR_Workshops} & Val & 10.9k & 5\\
Frame Interp & ~\cite{scharstein2002taxonomy} & Val & 1.9 & 5\\
Video Deblur & ~\cite{baker2011database} & Val & 9.4 & 5\\
Colorization & ~\cite{russakovsky2015imagenet} & Val & 4.7 & 5\\ \midrule
\textbf{2AFC--Real Alg [Val]} & -- & Val & 26.9k & 5 \\ \midrule \midrule
Traditional & ~\cite{dang2015raise} & Val & 4.8k & 3 \\
CNN-based & ~\cite{dang2015raise} & Val & 4.8k & 3 \\ \midrule
\textbf{JND--Distort} & -- & Val & 9.6k & 3 \\ \midrule \midrule
\vspace{-6mm}

\end{tabular}
}
\label{tab:dataset_split}
\caption{\textbf{Our dataset breakdown}. We split our 2AFC dataset in to three main portions (1,2) training and test sets with our distortions. Our training and test sets contain patches sampled from the MIT5k~\cite{fivek} and RAISE1k~\cite{dang2015raise} datasets, respectively  (3) a test set containing real algorithm outputs, containing patches from a variety of applications. Our JND data is on traditional and CNN-based distortions.}
\vspace{-8mm}
\end{center}
\end{table}

\paragraph{Just noticeable differences (JND)}

A potential shortcoming of the 2AFC task is that it is ``cognitively penetrable," in the sense that participants can consciously choose which respects of similarity they will choose to focus on in completing the task \cite{medin1993respects}, which introduces subjectivity into the judgments. To validate that the judgments actually reflected something objective and meaningful, we also collected user judgments of ``just noticeable differences" (JNDs). We show a reference image, followed by a randomly distorted image, and ask a human if the images are the same or different.
The two image patches are shown for 1 second each, with a 250 ms gap in between. 
Two images which look similar may be easily confused, and a good perceptual metric will be able to order pairs from most to least confusable. 
JND tests like this may be considered less subjective, since there is a single correct answer for each judgment, and participants are presumed to be aware of what correct behavior entails. We gather 3 JND observations for each of the 4.8k patches in our traditional and CNN-based validation sets. 
Each subject is shown 160 pairs, along with 40 sentinels (32 identical and 8 with large Gaussian noise distortion applied).
We also provide a short training period of 10 pairs which contain 4 ``same" pairs, 1 obviously different pair, and 5 ``different" pairs generated by our distortions. 
We chose to do this in order to prime the users towards expecting approximately $40\%$ of the patch pairs to be identical. 
Indeed, $36.4\%$ of the pairs were marked ``same" ($70.4\%$ of sentinels and $27.9\%$ of test pairs).

\begin{figure*}
  \centering
  \includegraphics[width=1.0\linewidth]{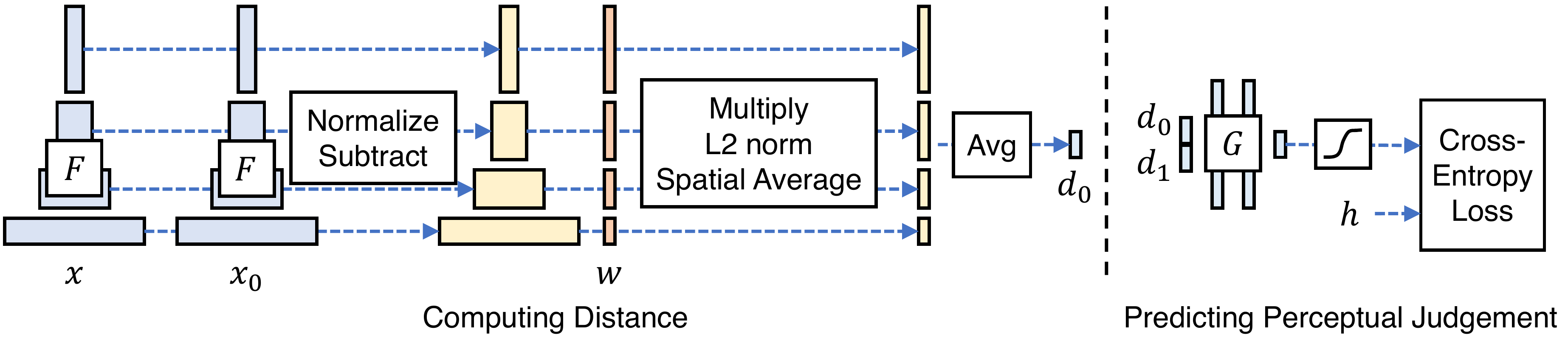}
\vspace{-6mm}
\caption{\textbf{Computing distance from a network} (Left) To compute a distance $d_0$ between two patches, $x$, $x_0$, given a network $\mathcal{F}$, we first compute deep embeddings, normalize the activations in the channel dimension, scale each channel by vector $w$, and take the $\ell_2$ distance. We then average across spatial dimension and across all layers. (Right) A small network $\mathcal{G}$ is trained to predict perceptual judgment $h$ from distance pair ($d_0,d_1$).}
\label{fig:network}
\vspace{-1mm}
\end{figure*}

\begin{figure*}[t]
\centering
\begin{tabular}{*{2}{c@{\hspace{3px}}}}
\textbf{Distortions} & \textbf{Real algorithms} \\
\includegraphics[width=.5\linewidth]{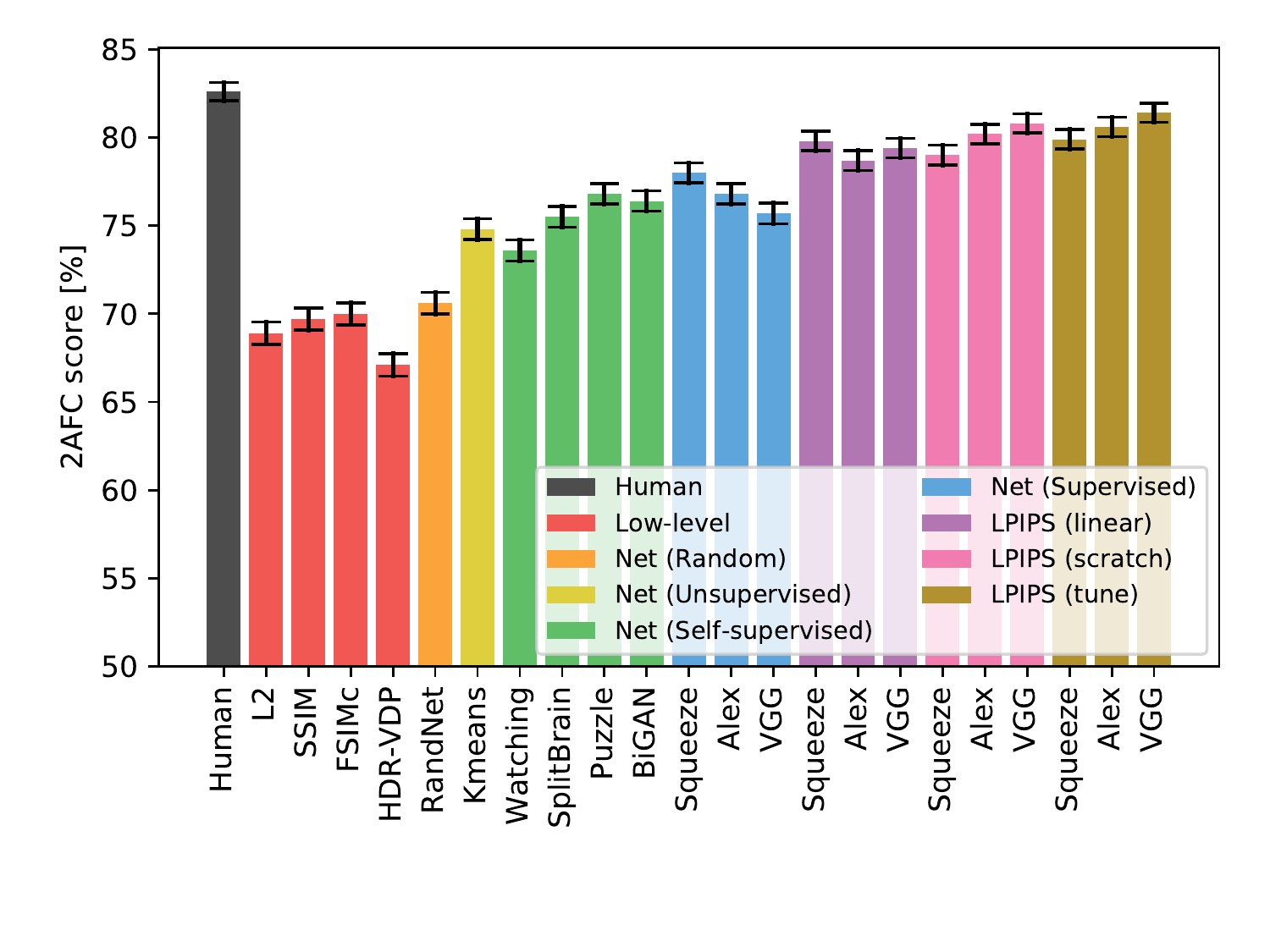} &
\includegraphics[width=.5\linewidth]{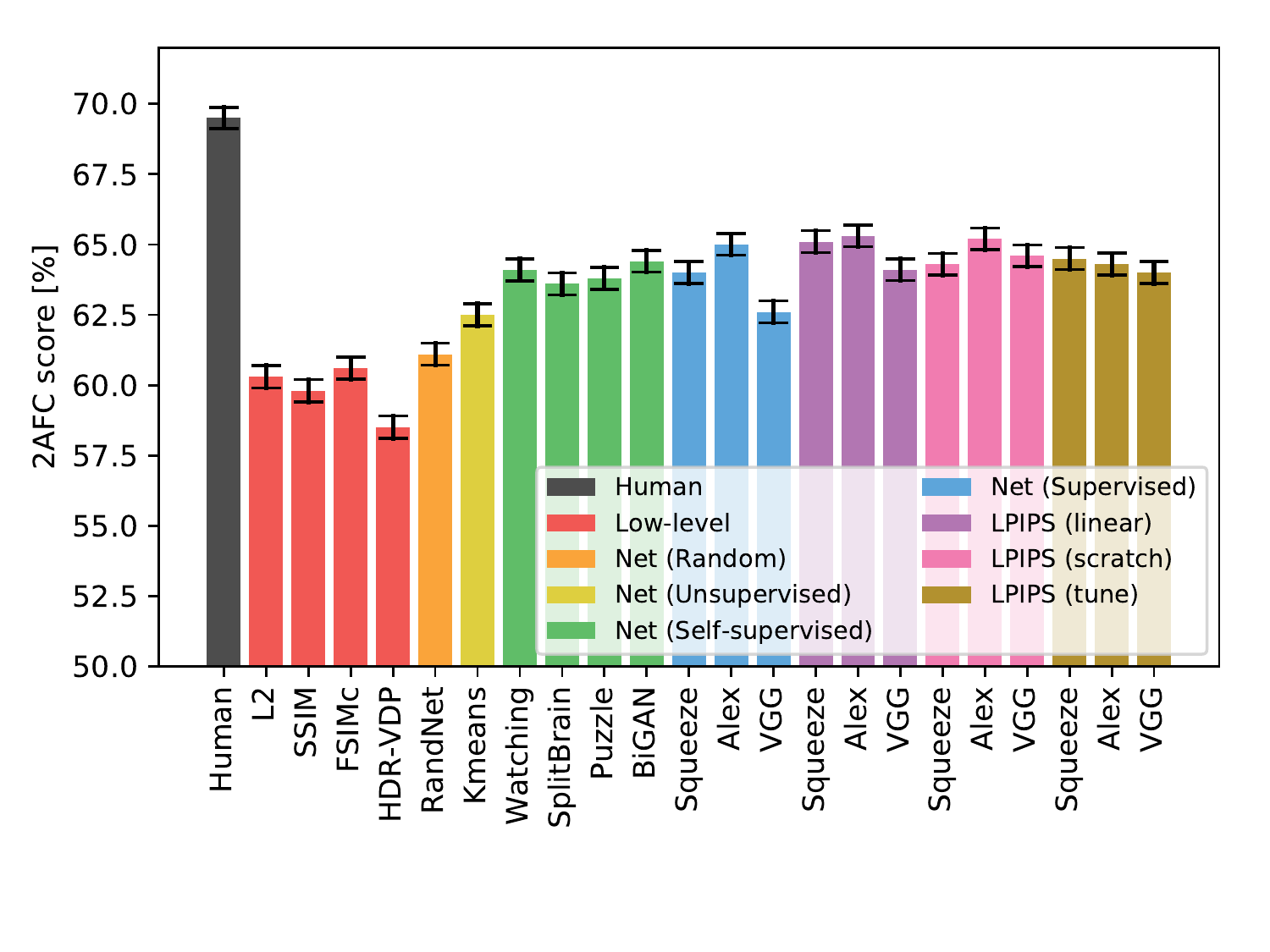} \\
\end{tabular}
\vspace{-12mm}
\caption{\label{fig:quant}
\textbf{Quantitative comparison.} We show a quantitative comparison across metrics on our test sets. (Left) Results averaged across our traditional and CNN-based distortions. (Right) Results averaged across our 4 real algorithm sets.}
\vspace{-2mm}
\end{figure*}

\section{Deep Feature Spaces}

We evaluate feature distances in different networks. For a given convolutional layer, we compute cosine distance (in the channel dimension) and average across spatial dimensions and layers of the network. We also discuss how to tune an existing network on our data.

\paragraph{Network architectures} We evaluate the SqueezeNet~\cite{iandola2016squeezenet}, AlexNet~\cite{krizhevsky2012imagenet}, and VGG~\cite{simonyan2014very} architectures. We use 5 \texttt{conv} layers from the VGG network, which has become the de facto standard for image generation tasks~\cite{gatys2016image,dosovitskiy2016generating,chen2017photographic}.
We also compare against the shallower AlexNet network, which may more closely match the architecture of the human visual cortex~\cite{yamins2016using}. 
We use the \texttt{conv1}-\texttt{conv5} layers from~\cite{krizhevsky2014one}. Finally, the SqueezeNet architecture was designed to be extremely lightweight ($2.8$ MB) in size, with similar classification performance to AlexNet. We use the first \texttt{conv} layer and some subsequent ``\texttt{fire}" modules.

We additionally evaluate self-supervised methods, including puzzle-solving~\cite{noroozi2016unsupervised}, cross-channel prediction~\cite{zhang2016colorful,zhang2017split}, learning from video~\cite{pathak2017learning}, and generative modeling~\cite{donahue2016adversarial}. We use publicly available networks from these and other methods, which use variants of AlexNet~\cite{krizhevsky2012imagenet}.

\paragraph{Network activations to distance} Figure \ref{fig:network} (left) and Equation~\ref{eqn:dist} illustrate how we obtain the distance between reference and distorted patches ${x,x_0}$ with network $\mathcal{F}$. We extract feature stack from $L$ layers and unit-normalize in the channel dimension, which we designate as $\hat{y}^l, \hat{y}_0^l \in \mathds{R}^{H_l\times W_l\times C_l}$ for layer $l$. We scale the activations channel-wise by vector $w^l \in \mathds{R}^{C_l}$ and compute the $\ell_2$ distance. Finally, we average spatially and sum channel-wise. Note that using $w_l=1 \forall l$ is equivalent to computing cosine distance.

\vspace{-4mm}
\begin{equation}
d(x,x_0) = \sum_l \dfrac{1}{H_l W_l} \sum_{h,w} || w_l \odot ( \hat{y}_{hw}^l - \hat{y}_{0hw}^l ) ||_2^2
\label{eqn:dist}
\end{equation}

\vspace{-2mm}

\paragraph{Training on our data} We consider a few variants for training with our perceptual judgments: \tbfit{lin}, \tbfit{tune}, and \tbfit{scratch}. For the \tbfit{lin} configuration, we keep pre-trained network weights $\mathcal{F}$ fixed, and learn linear weights $w$ on top. This constitutes a ``perceptual calibration" of a few parameters in an existing feature space. 
For example, for the VGG network, 1472 parameters are learned. For the \tbfit{tune} configuration, we initialize from a pre-trained classification model, and allow all the weights for network $\mathcal{F}$ to be fine-tuned. Finally, for \tbfit{scratch}, we initialize the network from random Gaussian weights and train it entirely on our judgments. Overall, we refer to these as variants of our proposed \textbf{Learned Perceptual Image Patch Similarity (LPIPS)} metric. We illustrate the training loss function in Figure~\ref{fig:network} (right) and describe it further in the appendix.

\section{Experiments}

Results on our validation sets are shown in Figure~\ref{fig:quant}. We first evaluate how well our metrics and networks work. All validation sets contain 5 pairwise judgments for each triplet. Because this is an inherently noisy process, we compute agreement of an algorithm with \textit{all} of the judgments. For example, if there are 4 preferences for $x_0$ and 1 for $x_1$, an algorithm which predicts the more popular choice $x_0$ would receive $80\%$ credit. If a given example is scored with fraction $p$ humans in one direction and $1-p$ in the other, a human would achieve score $p^2+(1-p)^2$ on expectation.

\subsection{Evaluations}

\begin{figure}[t]
  \centering
  \includegraphics[width=1.\linewidth]{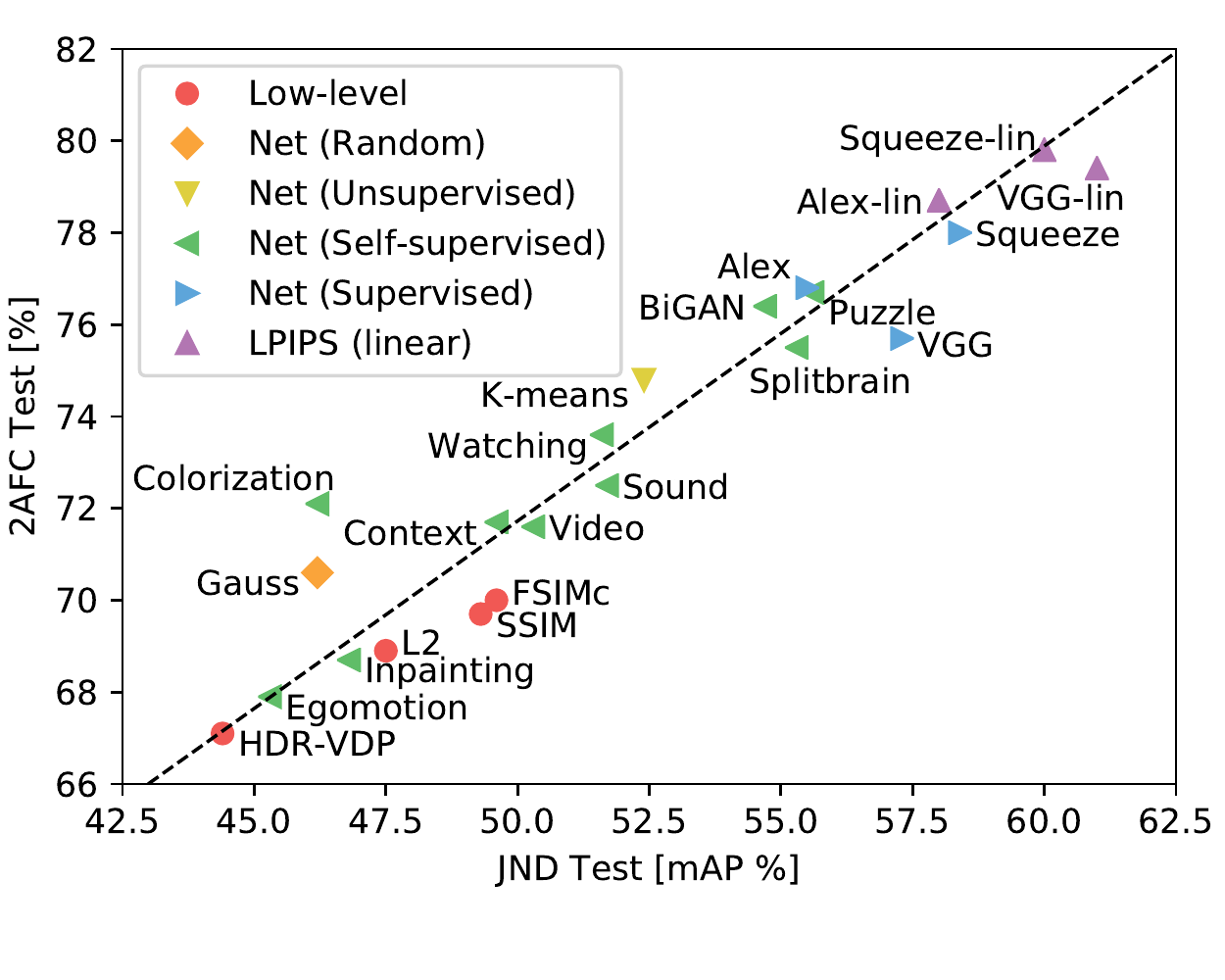}
\vspace{-12mm}
\caption{\textbf{Correlating Perceptual Tests.} We show performance across methods, including unsupervised~\cite{krahenbuhl2015data}, self-supervised~\cite{agrawal2015learning,pathakCVPR16context,doersch2015unsupervised,wang2015unsupervised,zhang2016colorful,owens2016visually,pathak2017learning,noroozi2016unsupervised,donahue2016adversarial,zhang2017split}, supervised~\cite{krizhevsky2014one,simonyan2014very,iandola2016squeezenet}, and our perceptually-learned metrics (LPIPS). The scores are on our 2AFC and JND tests, averaged across traditional and CNN-based distortions.}
\label{fig:trip_vs_jnd}
\vspace{-2mm}
\end{figure}

\paragraph{How well do low-level metrics and classification networks perform?} Figure~\ref{fig:quant} shows the performance of various low-level metrics (in red), deep networks, and human ceiling (in black). The scores are averaged across the 2 distortion test sets (traditional+CNN-based) in Figure~\ref{fig:quant} (left), and 4 real algorithm benchmarks (superresolution, frame interpolation, video deblurring, colorization) in Figure~\ref{fig:quant} (right). All scores within each test set are shown in the appendix. Averaged across all 6 test sets, humans are $73.9\%$ consistent. Interestingly, the supervised networks perform at about the same level to each other, at $68.6\%$, $68.9\%$, and $67.0\%$, even across variation in model sizes -- SqueezeNet ($2.8$ MB), AlexNet ($9.1$ MB), and VGG ($58.9$ MB) (only convolutional layers are counted). They all perform better than traditional metrics $\ell_2$, SSIM, and FSIM at $63.2\%$, $63.1\%$, $63.8\%$, respectively. Despite its common use, SSIM was not designed for situations where geometric distortion is a large factor~\cite{sampat2009complex}.

\paragraph{Does the network have to be trained on classification?} In Figure~\ref{fig:quant}, we show model performance across a variety of unsupervised and self-supervised tasks, shown in green -- generative modeling with BiGANs~\cite{donahue2016adversarial}, solving puzzles~\cite{noroozi2016unsupervised}, cross-channel prediction~\cite{zhang2017split}, and segmenting foreground objects from video~\cite{pathak2017learning}. These self-supervised tasks perform on par with classification networks. This indicates that tasks across a large spectrum can induce representations which transfer well to perceptual distances. Also, the performance of the stacked k-means method~\cite{krahenbuhl2015data}, shown in yellow, outperforms low-level metrics. Random networks, shown in orange, with weights drawn from a Gaussian, do not yield much improvement. This indicates that the combination of network structure, along with orienting filters in directions where data is more dense, can better correlate to perceptual judgments.

\begin{table}[t!]
\begin{center}
\resizebox{\linewidth}{!} {
\begin{tabular}{ c c | c c c c | c}
& & \textbf{2AFC} & \textbf{JND} & \textbf{Class.} & \textbf{Det.} & \textbf{Avg} \\
\midrule
\textbf{Perceptual} & \textbf{2AFC} & -- & .928 & .640 & .363 & .644 \\
\textbf{Perceptual} & \textbf{JND} & .928 & -- & .612 & .232 & .591 \\
\textbf{PASCAL} & \textbf{Classification} & .640 & .612 & -- & .429 & .560 \\
\textbf{PASCAL} & \textbf{Detection} &.363 & .232 & .429 & -- & .341 \\
\end{tabular}
}
\vspace{-2mm}
\caption{\textbf{Task correlation}. We correlate scores between our low-level perceptual tests along with high-level semantic tests across methods. Perceptual scores are averaged between traditional and CNN-based distortion sets. Correlation scores are computed for AlexNet-like architectures.}
\label{tab:corr}
\vspace{-8mm}
\end{center}
\end{table}

In Table~\ref{fig:trip_vs_jnd}, we explore how well our perceptual task correlates to semantic tasks on the PASCAL dataset~\cite{pascal-voc-2007}, using results summarized in~\cite{zhang2017split}, including additional self-supervised methods~\cite{agrawal2015learning,pathakCVPR16context,doersch2015unsupervised,wang2015unsupervised,zhang2016colorful,owens2016visually}. We compute the correlation coefficient between each task (perceptual or semantic) across different methods. The correlation from our 2AFC distortion preference task to classification and detection is .640 and .363, respectively. Interestingly, this is similar to the correlation between the classification and detection tasks (.429), even though both are considered ``high-level" semantic tasks, and our perceptual task is ``low-level."

\begin{figure*}
\centering
\begin{subfigure}{1.\textwidth}
  \includegraphics[width=1.\linewidth]{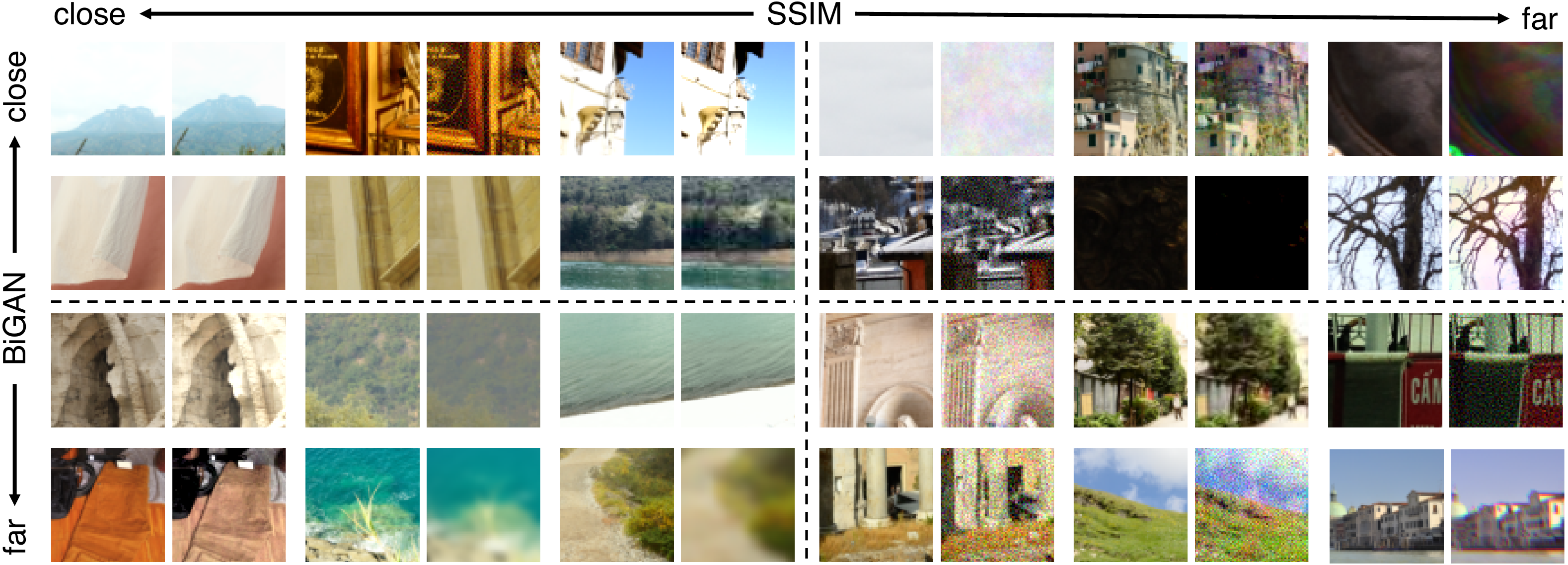}
  \vspace{-5mm}
\end{subfigure}
\caption{\textbf{Qualitative comparisons on distortions.} We show qualitative comparison on traditional distortions, using the SSIM~\cite{wang2004image} metric and BiGAN network~\cite{donahue2016adversarial}. We show examples where the metrics agree and disagree. A primary difference is that deep embeddings appear to be more sensitive to blur. Please see the appendix for additional examples.}
\label{fig:qual_comp}
\vspace{-4mm}
\end{figure*}

\paragraph{Do metrics correlate across different perceptual tasks?} We test if training for the 2AFC distortion preference test corresponds with another perceptual task, the JND test. We order patch pairs by ascending order by a given metric, and compute precision-recall on our CNN-based distortions -- for a good metric, patches which are close together are more likely to be confused for being the same. We compute area under the curve, known as mAP~\cite{pascal-voc-2007}. The 2AFC distortion preference test has high correlation to JND: $\rho=.928$ when averaging the results across distortion types. Figure~\ref{fig:trip_vs_jnd} shows how different methods perform under each perceptual test. This indicates that 2AFC generalizes to another perceptual test and is giving us signal regarding human judgments.

\paragraph{Can we train a metric on traditional and CNN-based distortions?} In Figure~\ref{fig:quant}, we show performance using our \tbfit{lin}, \tbfit{scratch}, and \tbfit{tune} configurations, shown in purple, pink, and brown, respectively. When validating on the traditional and CNN-based distortions (Figure~\ref{fig:quant}(a)), we see improvements. Allowing the network to tune all the way through (brown) achieves higher performance than simply learning linear weights (purple) or training from scratch (pink). The higher capacity network VGG also performs better than the lower capacity SqueezeNet and AlexNet architectures. These results verify that networks can indeed learn from perceptual judgments.

\paragraph{Does training on traditional and CNN-based distortions transfer to real-world scenarios?} We are more interested in how performance generalizes to \textit{real-world algorithms}, shown in Figure~\ref{fig:quant}(b). The SqueezeNet, AlexNet, and VGG architectures start at $64.0\%$, $65.0\%$, and $62.6\%$, respectively. Learning a linear classifier (purple) improves performance for all networks. Across the 3 networks and 4 real-algorithm tasks, 11 of the 12 scores improved, indicating that ``calibrating" activations on a pre-existing representation using our data is a safe way to achieve a small boost in performance ($1.1\%$, $0.3\%$, and $1.5\%$, respectively). Training a network from scratch (pink) yields slightly lower performance for AlexNet, and slightly higher performance for VGG than linear calibration. However, these still outperform low-level metrics. This indicates that the distortions we have expressed do project onto our test-time tasks of judging real algorithms. 

Interestingly, starting with a pre-trained network and tuning throughout \textit{lowers} transfer performance. This is an interesting negative result, as training for a low-level perceptual task directly does not necessarily perform as well as transferring a representation trained for the high-level task.

\paragraph{Where do deep metrics and low-level metrics disagree?} In Figure~\ref{fig:qual_comp}, we show a qualitative comparison across our traditional distortions for a deep method, BiGANs~\cite{donahue2016adversarial}, and a representation traditional perceptual method, SSIM~\cite{wang2004image}. Pairs which BiGAN perceives to be far but SSIM to be close generally contain some blur. BiGAN tends to perceive correlated noise patterns to be a smaller distortion than SSIM.

\section{Conclusions}
\vspace{-1mm}

Our results indicate that networks trained to solve challenging visual prediction and modeling tasks end up learning a representation of the world that correlates well with perceptual judgments. A similar story has recently emerged in the representation learning literature: networks trained on self-supervised and unsupervised objectives end up learning a representation that is also effective at semantic tasks~\cite{doersch2015unsupervised}. Interestingly, recent findings in neuroscience make much the same point: representations trained on computer vision tasks also end up being effective models of neural activity in macaque visual cortex~\cite{yamins2016using}. Moreover (and roughly speaking), the stronger the representation is at the computer vision task, the stronger it is as a model of cortical activity. Our paper makes a similar finding: the stronger a feature set is at classification and detection, the stronger it is as a model of perceptual similarity judgments, as suggested in Table~\ref{tab:corr}. Together, these results suggest that a good feature is a good feature. Features that are good at semantic tasks are also good at self-supervised and unsupervised tasks, and also provide good models of both human perceptual behavior and macaque neural activity. This last point aligns with the ``rational analysis" explanation of visual cognition~\cite{anderson1990adaptive}, suggesting that the idiosyncrasies of biological perception arise as a consequence of a rational agent attempting to solve natural tasks. Further refining the degree to which this is true is an important question for future research.

\vspace{-1mm}
{\paragraph{Acknowledgements} This research was supported, in part, by grants from Berkeley Deep Drive, NSF IIS-1633310, and hardware donations by NVIDIA. We thank members of the Berkeley AI Research Lab and Adobe Research for helpful discussions. We thank Alan Bovik for his insightful comments. We also thank Radu Timofte, Zhaowen Wang, Michael Waechter, Simon Niklaus, and Sergio Guadarrama for help preparing data. RZ is partially supported by an Adobe Research Fellowship and much of this work was done while RZ was an intern at Adobe Research.}

{\small
\bibliographystyle{ieee}
\bibliography{egbib}
}

\appendix
\section*{Appendix}

We show full quantitative details in Appendix \ref{sec:quant}. We also discuss training details in Appendix \ref{sec:train}. Finally, we show results on the TID2013 dataset~\cite{ponomarenko2015image} in Appendix \ref{sec:tid}.

\section{Quantitative Results}
\label{sec:quant}

\begin{figure*}[t]
\centering
\begin{tabular}{*{2}{c@{\hspace{3px}}}}
\textbf{Distortions (Traditional)} & \textbf{Distortions (CNN-Based)} \\
\includegraphics[width=.5\linewidth]{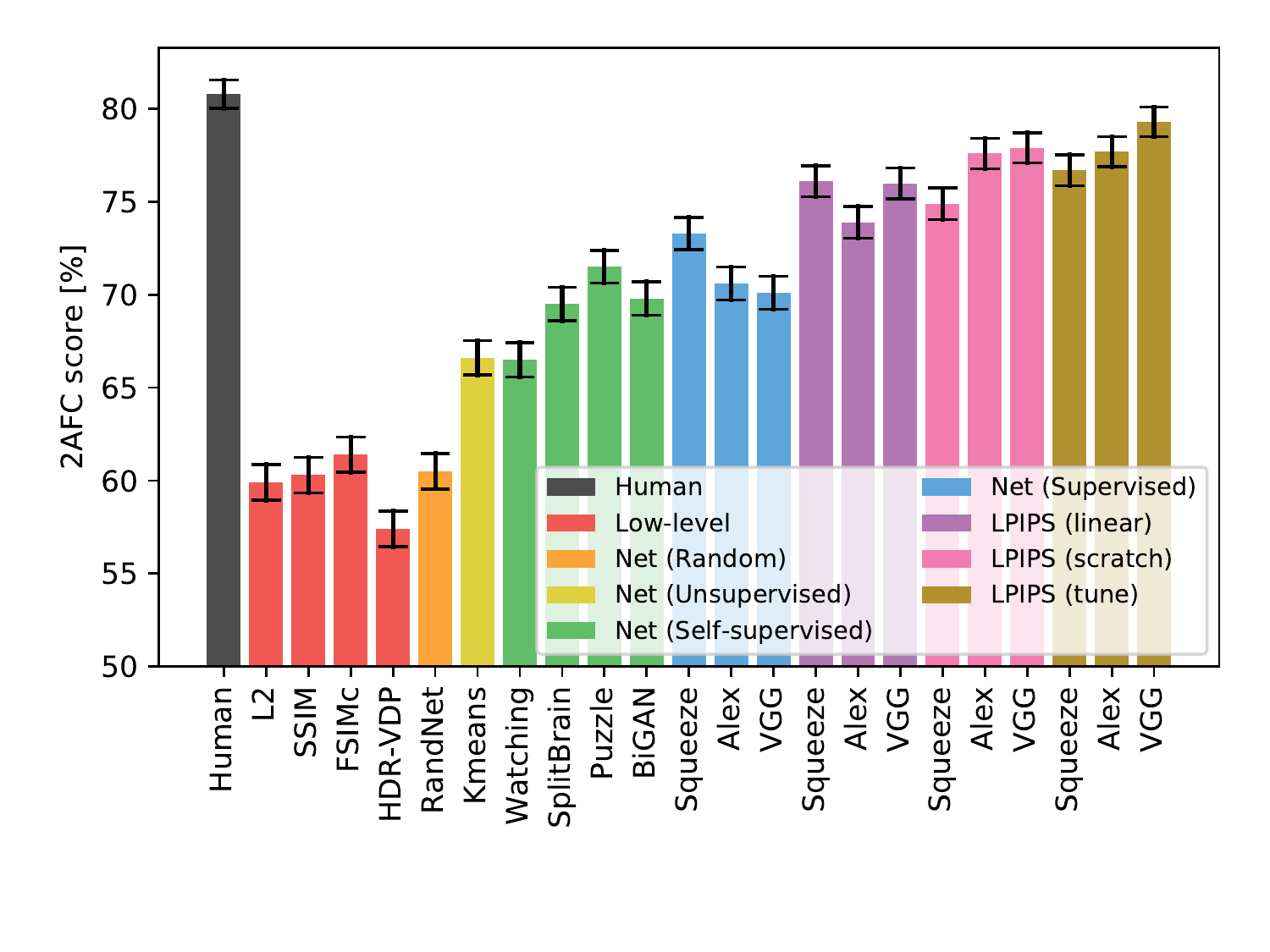}
&
\includegraphics[width=.5\linewidth]{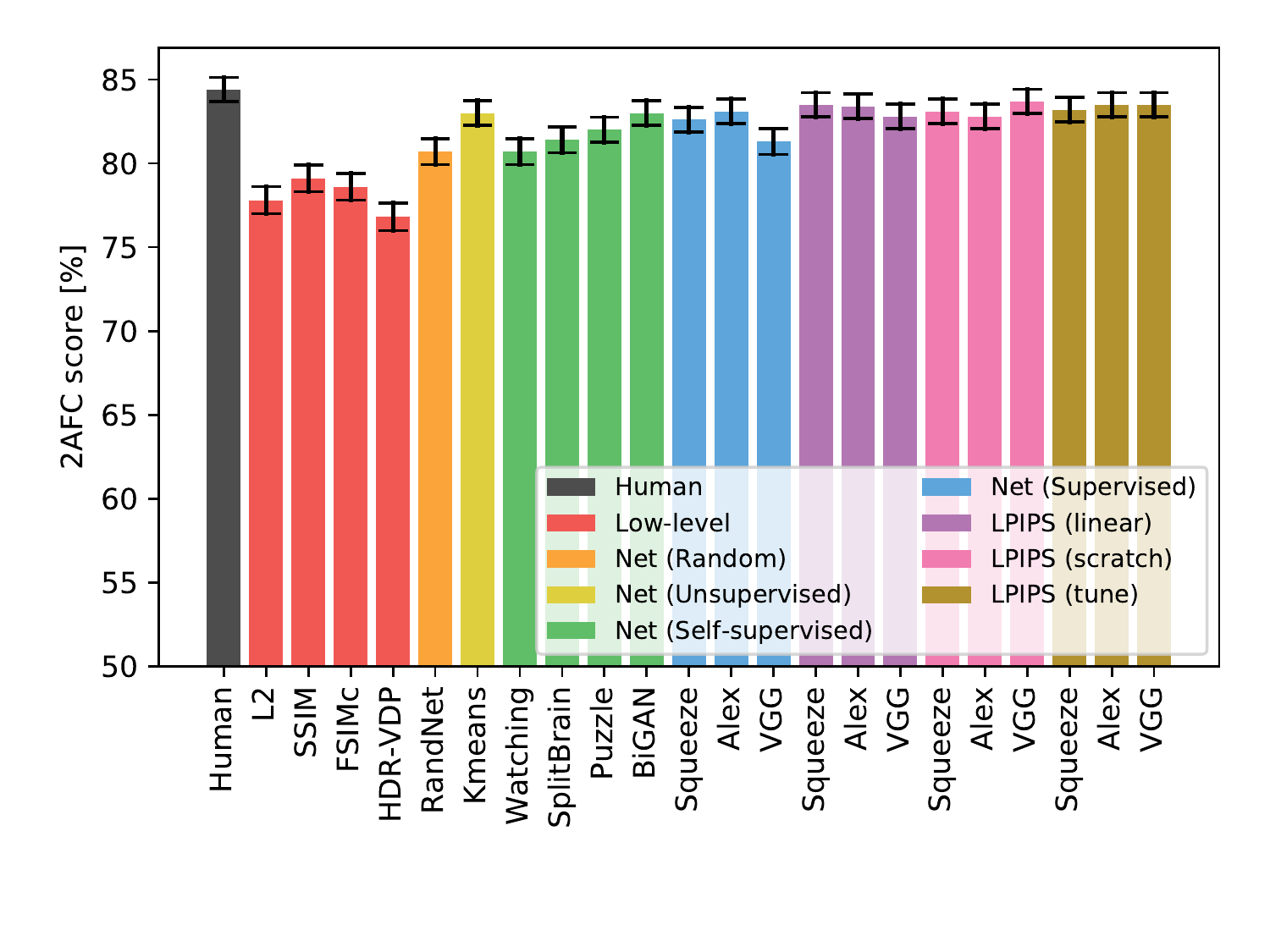} \\
\end{tabular}
\vspace{-10mm}
\caption{\textbf{Individual results} (left) traditional distortions (right) CNN-based distortions}
\label{fig:quant1}
\end{figure*}

\begin{figure*}[t]
\centering
\begin{tabular}{*{2}{c@{\hspace{3px}}}}
\textbf{Real Algorithms (Superresolution)} & \textbf{Real Algorithms (Frame Interpolation)} \\
\includegraphics[width=.5\linewidth]{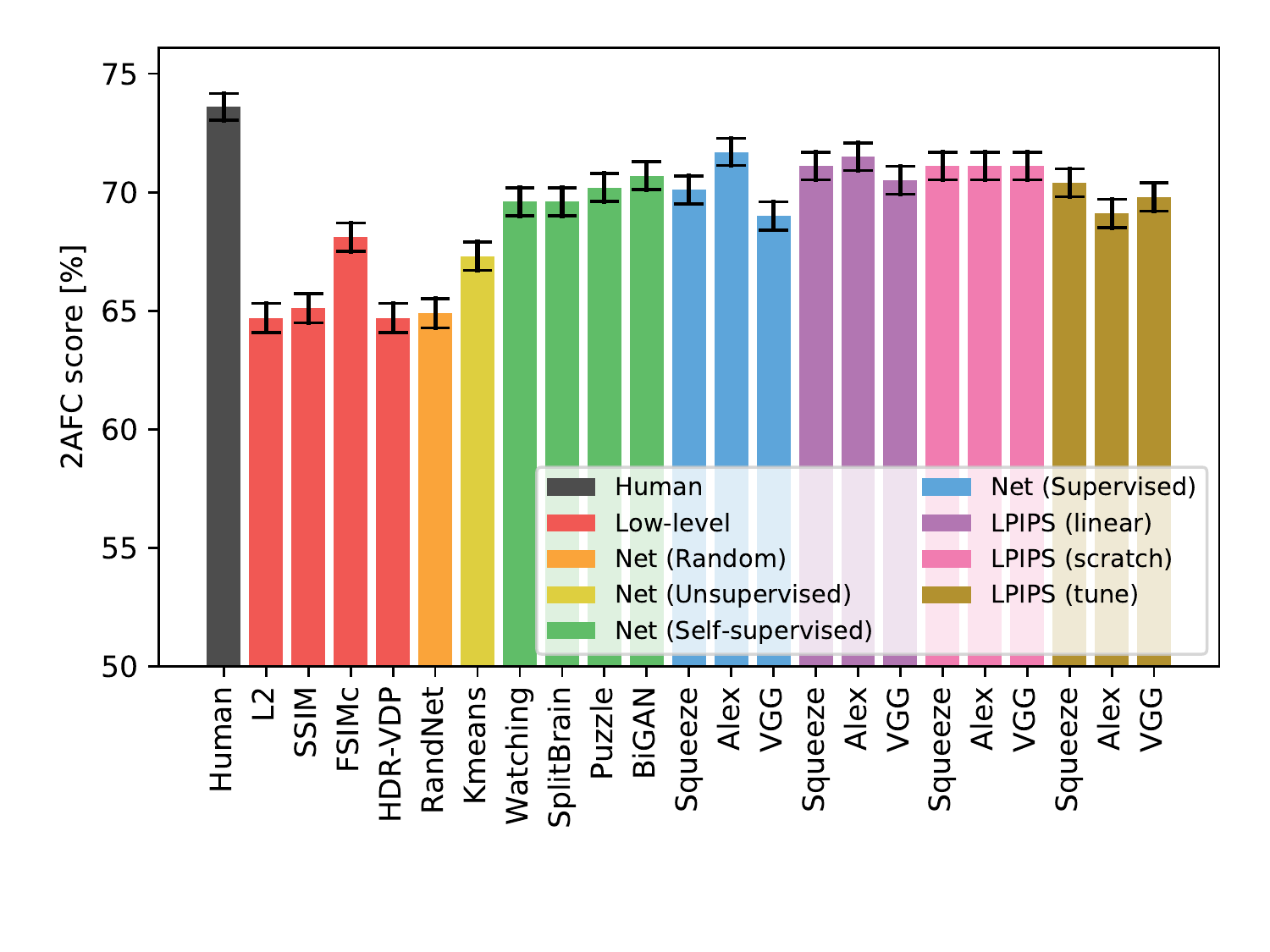} &
\includegraphics[width=.5\linewidth]{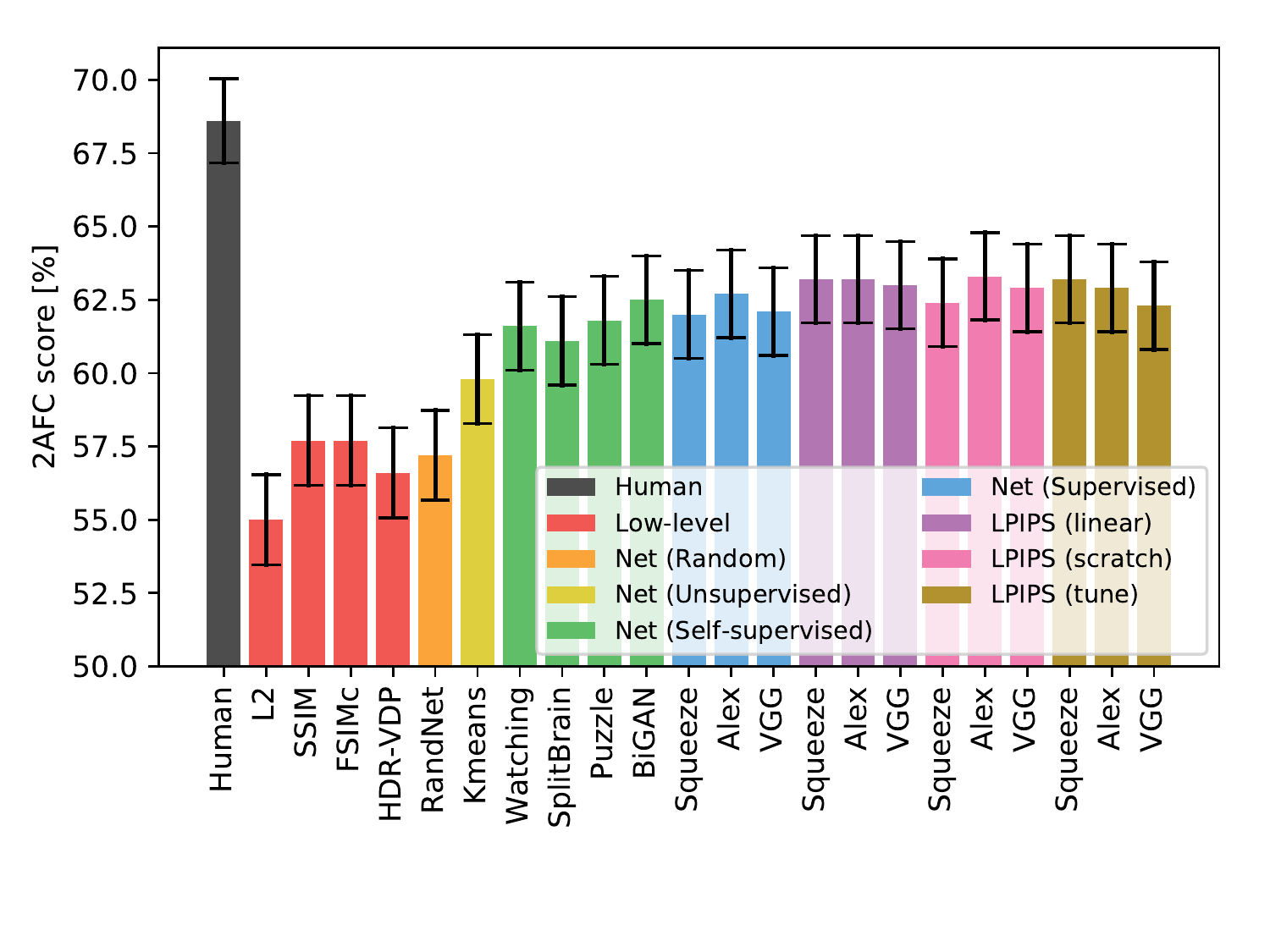} \\
\end{tabular}
\vspace{-10mm}
\caption{\textbf{Individual results} (left) superresolution (right) frame interpolation}
\label{fig:quant2}
\end{figure*}

\begin{figure*}[h!]
\centering
\begin{tabular}{*{2}{c@{\hspace{3px}}}}
\textbf{Real Algorithms (Video Deblurring)} & \textbf{Real Algorithms (Colorization)} \\
\includegraphics[width=.5\linewidth]{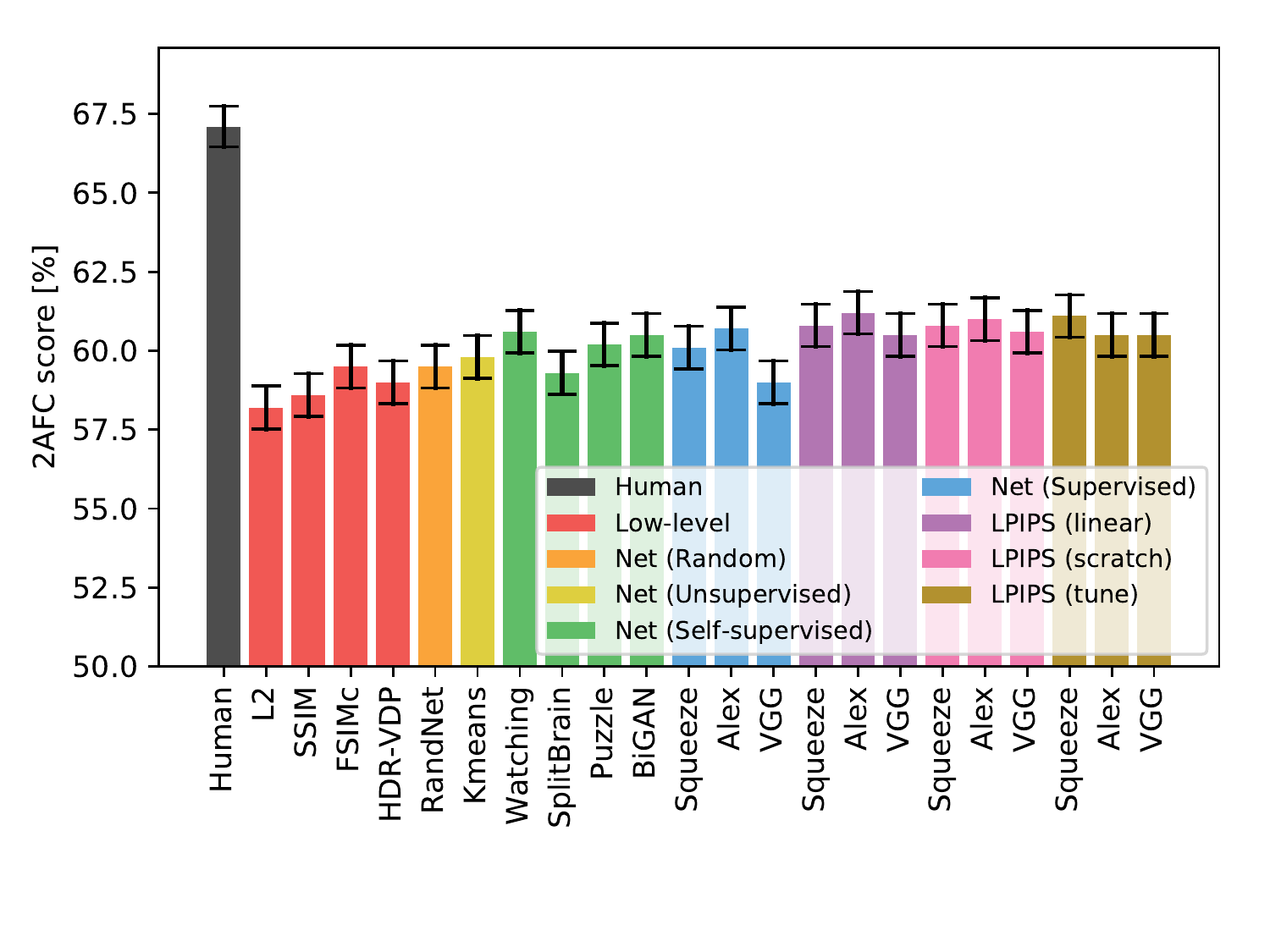} &
\includegraphics[width=.5\linewidth]{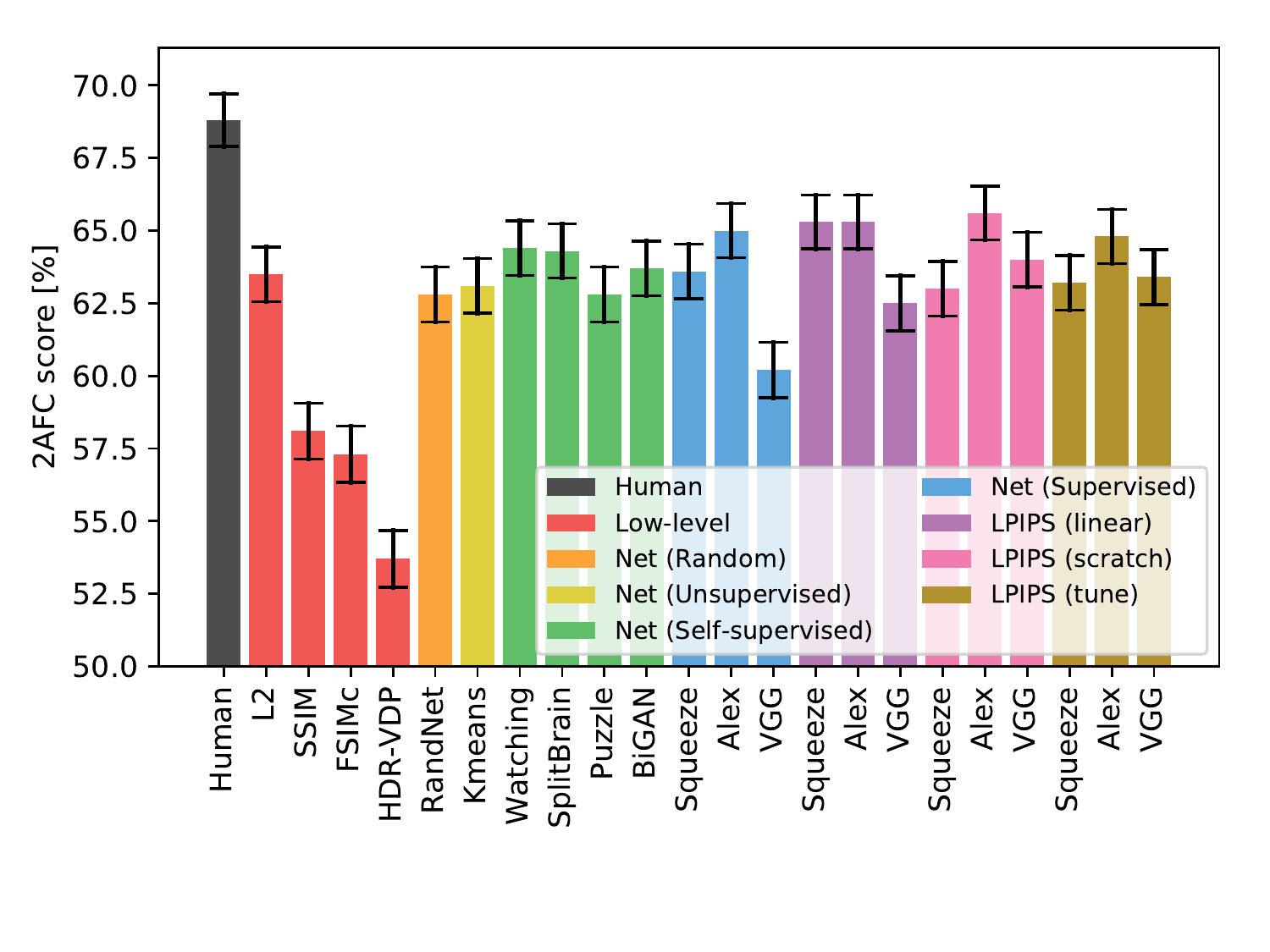} \\
\end{tabular}
\vspace{-10mm}
\caption{\textbf{Individual results} (left) video deblurring (right) colorization}
\label{fig:quant3}
\end{figure*}

\begin{table*}[t]
\begin{center}
\resizebox{\linewidth}{!} {
\begin{tabular}{l l@{\hskip .05in} c@{\hskip .05in} c@{\hskip .05in} c@{\hskip .15in} c@{\hskip .05in} @{\hskip .05in}c@{\hskip .05in} c@{\hskip .05in} c@{\hskip .05in} c@{\hskip .15in} c@{\hskip .05in} } \toprule

\multirow{3}[3]{*}{\textbf{Subtype}} &\multirow{3}[3]{*}{\textbf{Metric}} & \multicolumn{3}{c}{\textbf{Distortions}} & \multicolumn{5}{c}{\textbf{Real Algorithms}} & \multirow{1}{*}{\textbf{All}} \\ \cmidrule(lr){3-5} \cmidrule(lr){6-10} \cmidrule(lr){11-11} 

& & \textbf{Trad-} & \textbf{CNN-} & \multirow{2}{*}{\textbf{All}} & \textbf{Super-} & \textbf{Video} & \textbf{Color-} & \textbf{Frame} & \multirow{2}{*}{\textbf{All}} & \multirow{2}{*}{\textbf{All}} \\
 
& & \textbf{itional} & \textbf{Based} & \textbf{} & \textbf{res} & \textbf{Deblur} & \textbf{ization} & \textbf{Interp} & \\ \midrule

Oracle & Human & 80.8 & 84.4 & 82.6 & 73.4 & 67.1 & 68.8 & 68.6 & 69.5 & 73.9 \\ \midrule

\multirow{4}{*}{Low-level} & L2& 59.9 & 77.8 & 68.9 & 64.7 & 58.2 & 63.5 & 55.0 & 60.3 & 63.2 \\
& SSIM~\cite{wang2004image}& 60.3 & 79.1 & 69.7 & 65.1 & 58.6 & 58.1 & 57.7 & 59.8 & 63.1 \\
& FSIMc~\cite{zhang2011fsim}& 61.4 & 78.6 & 70.0 & 68.1 & 59.5 & 57.3 & 57.7 & 60.6 & 63.8 \\
& HDR-VDP~\cite{mantiuk2011hdr}& 57.4 & 76.8 & 67.1 & 64.7 & 59.0 & 53.7 & 56.6 & 58.5 & 61.4 \\ \midrule
Net (Random) & Gaussian & 60.5 & 80.7 & 70.6 & 64.9 & 59.5 & 62.8 & 57.2 & 61.1 & 64.3 \\ \midrule
Net (Unsupervised) & K-means~\cite{krahenbuhl2015data}& 66.6 & \tbfit{83.0} & 74.8 & 67.3 & 59.8 & 63.1 & 59.8 & 62.5 & 66.6 \\ \midrule
\multirow{4}{*}{Net (Self-supervised)} & Watching~\cite{pathak2017learning}& 66.5 & 80.7 & 73.6 & 69.6 & 60.6 & 64.4 & 61.6 & 64.1 & 67.2 \\
& Split-Brain~\cite{zhang2017split}& 69.5 & 81.4 & 75.5 & 69.6 & 59.3 & 64.3 & 61.1 & 63.6 & 67.5 \\
& Puzzle~\cite{noroozi2016unsupervised}& 71.5 & 82.0 & 76.8 & 70.2 & 60.2 & 62.8 & 61.8 & 63.8 & 68.1 \\
& BiGAN~\cite{donahue2016adversarial}& 69.8 & \tbfit{83.0} & 76.4 & 70.7 & 60.5 & 63.7 & 62.5 & 64.4 & \tbfit{68.4} \\ \midrule
\multirow{3}{*}{Net (Supervised)} & SqueezeNet~\cite{iandola2016squeezenet}& \tbfu{73.3} & \tbfit{82.6} & \tbfu{78.0} & 70.1 & 60.1 & 63.6 & 62.0 & 64.0 & \tbfit{68.6} \\
& AlexNet~\cite{krizhevsky2014one}& 70.6 & \tbfu{83.1} & 76.8 & \tbfu{71.7} & 60.7 & 65.0 & 62.7 & \tbfit{65.0} & \tbfu{68.9} \\
& VGG~\cite{simonyan2014very}& 70.1 & 81.3 & 75.7 & 69.0 & 59.0 & 60.2 & 62.1 & 62.6 & 67.0 \\ \midrule
\multirow{7}{*}{*LPIPS (Learned} & Squeeze -- lin & \gray{76.1} & \gray{83.5} & \gray{79.8} & 71.1 & \tbfit{60.8} & \tbfit{65.3} & \tbfit{63.2} & \tbfit{65.1} & \gray{70.0} \\
\multirow{7}{*}{Perceptual Image} & Alex -- lin & \gray{73.9} & \gray{83.4} & \gray{78.7} & \tbfit{71.5} & \tbfu{61.2} & \tbfit{65.3} & \tbfit{63.2} & \tbfu{65.3} & \gray{69.8} \\
\multirow{7}{*}{Patch Similarity)} & VGG -- lin & \gray{76.0} & \gray{82.8} & \gray{79.4} & 70.5 & 60.5 & 62.5 & \tbfit{63.0} & 64.1 & \gray{69.2} \\ \cdashline{2-11}
& Squeeze -- scratch & \gray{74.9} & \gray{83.1} & \gray{79.0} & 71.1 & \tbfit{60.8} & 63.0 & 62.4 & 64.3 & \gray{69.2} \\
& Alex -- scratch & \gray{77.6} & \gray{82.8} & \gray{80.2} & 71.1 & \tbfit{61.0} & \tbfu{65.6} & \tbfu{63.3} & \tbfit{65.2} & \gray{\textbf{70.2}} \\
& VGG -- scratch & \gray{77.9} & \gray{\textbf{83.7}} & \gray{80.8} & 71.1 & 60.6 & 64.0 & \tbfit{62.9} & 64.6 & \gray{70.0} \\ \cdashline{2-11}
& Squeeze -- tune & \gray{76.7} & \gray{83.2} & \gray{79.9} & 70.4 & \tbfit{61.1} & 63.2 & \tbfit{63.2} & 64.5 & \gray{69.6} \\
& Alex -- tune & \gray{77.7} & \gray{83.5} & \gray{80.6} & 69.1 & 60.5 & 64.8 & \tbfit{62.9} & 64.3 & \gray{69.7} \\
& VGG -- tune & \gray{\textbf{79.3}} & \gray{83.5} & \gray{\textbf{81.4}} & 69.8 & 60.5 & 63.4 & 62.3 & 64.0 & \gray{69.8} \\

\bottomrule
\end{tabular}
}
\caption{\textbf{Results}. We show 2AFC scores (higher is better) across a spectrum of methods and test sets. The \tbfu{bolded \& underlined} values are the highest performing. The \tbfit{bolded \& italicized} values are within 0.5\% of highest. *LPIPS metrics are trained on the same traditional and CNN-based distortions, and as such have an advantage relative to other methods when testing on those same distortion types, even on unseen test images. These values are indicated by \gray{gray} values. The best gray value per column is also \gray{\textbf{bolded}}.}
\label{tab:res_quant}
\end{center}
\end{table*}

\begin{figure*}
\centering
\begin{subfigure}{1.\textwidth}
  \centering
  \includegraphics[width=1.\linewidth]{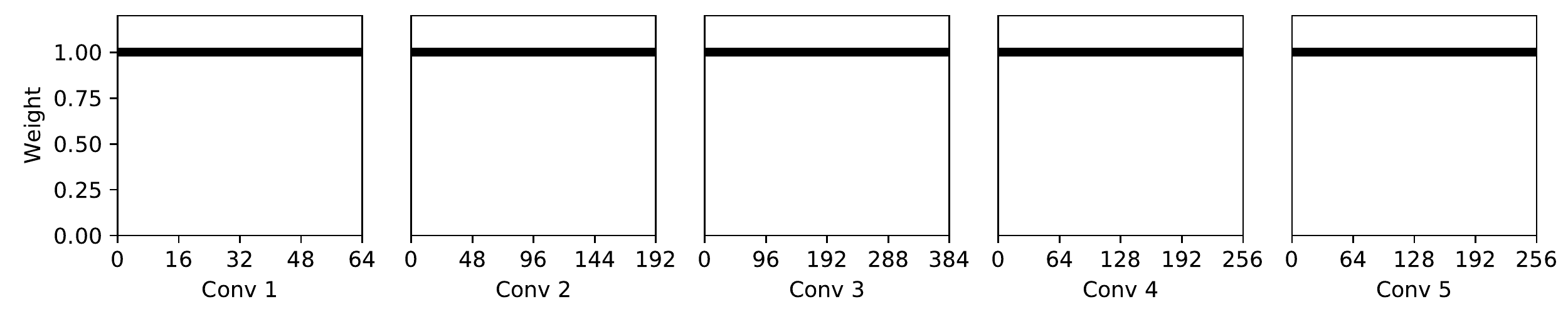}
  \vspace{-6mm}
  \caption{\textbf{Unlearned weights for AlexNet model (cosine distance)}}
  \label{fig:weights_ones}
\end{subfigure}

\begin{subfigure}{1.\textwidth}
  \centering
    \includegraphics[width=1.\linewidth]{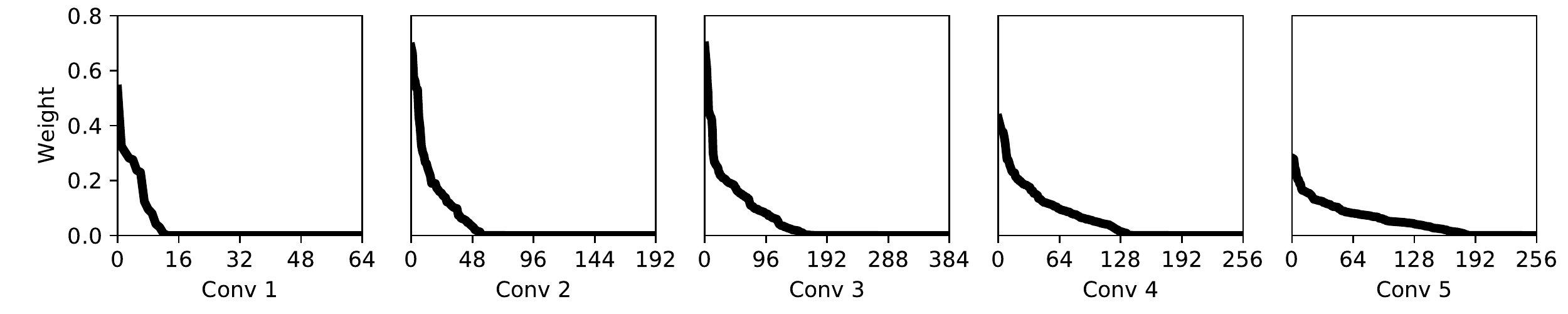}
    \vspace{-6mm}
    \caption{\textbf{Learned weights from \textit{Alex--lin} model}}
\label{fig:weights_learned}
\end{subfigure}

\begin{subfigure}{1.\textwidth}
  \centering
    \includegraphics[width=1.\linewidth]{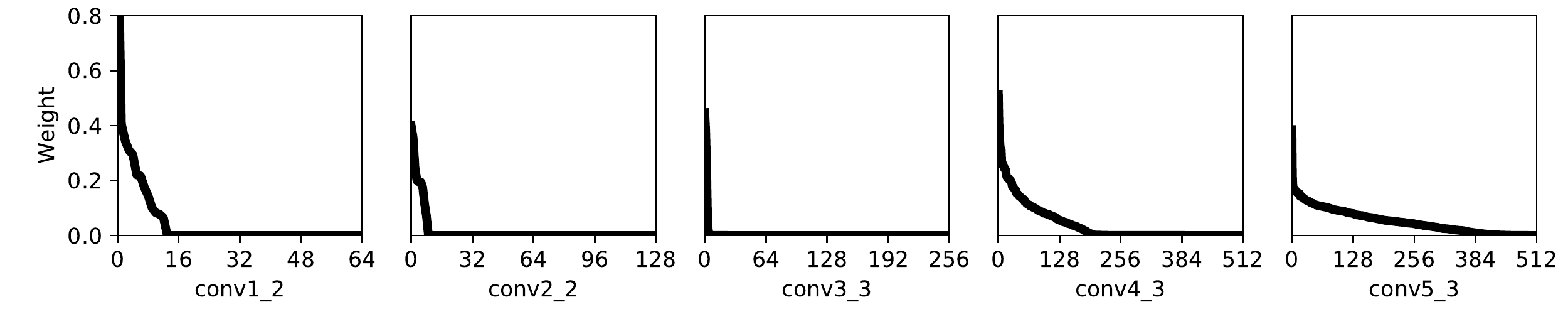}
    \vspace{-6mm}
    \caption{\textbf{Learned weights from \textit{VGG--lin} model}}
\label{fig:weights_learned_vgg}
\end{subfigure}

\begin{subfigure}{1.\textwidth}
  \centering
    \includegraphics[width=1.\linewidth]{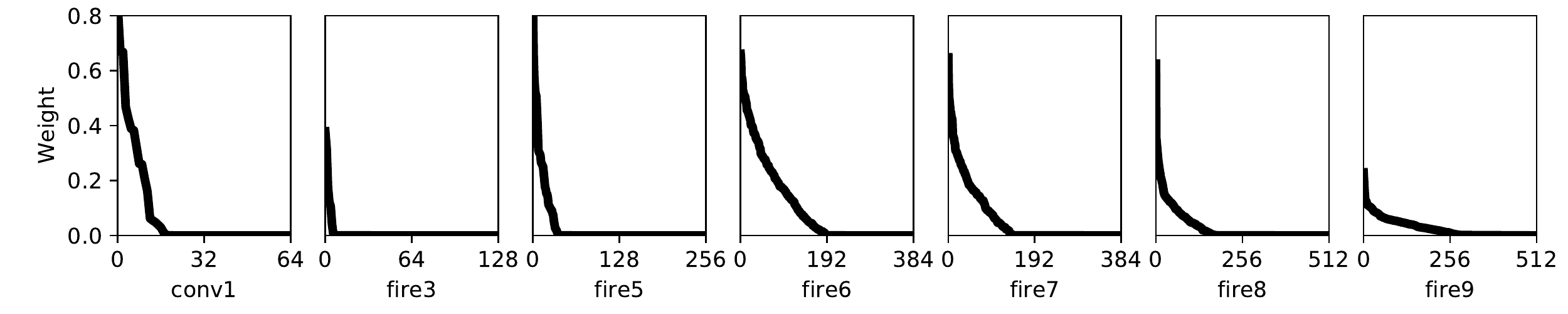}
    \vspace{-6mm}
    \caption{\textbf{Learned weights from \textit{Squeeze--lin} model}}
\label{fig:weights_learned_squeeze}
\end{subfigure}

\vspace{-2mm}
\caption{\textbf{Learned linear weights by layer.} (a) Unlearned weights correspond to using weighting 1 for each channel in each layer, which results in computing cosine distance. (b) We show the learned weights from each layer of our \textit{Alex--lin} model. This is the $w$ term in Figure~\ref{fig:network}. Each subplot shows the channel weights from each layer, sorted in descending order. The x-axis shows the channel number, and y-axis shows the weight. Weights are restricted to be non-negative, as image patches should not have negative distance. (c,d) Same as (b), but with the \textit{VGG--lin} and \textit{Squeeze--lin} models.}
\label{fig:weights}
\vspace{-4mm}
\end{figure*}

\begin{figure*}
\centering
\begin{subfigure}{1.\textwidth}
  \includegraphics[width=1.\linewidth]{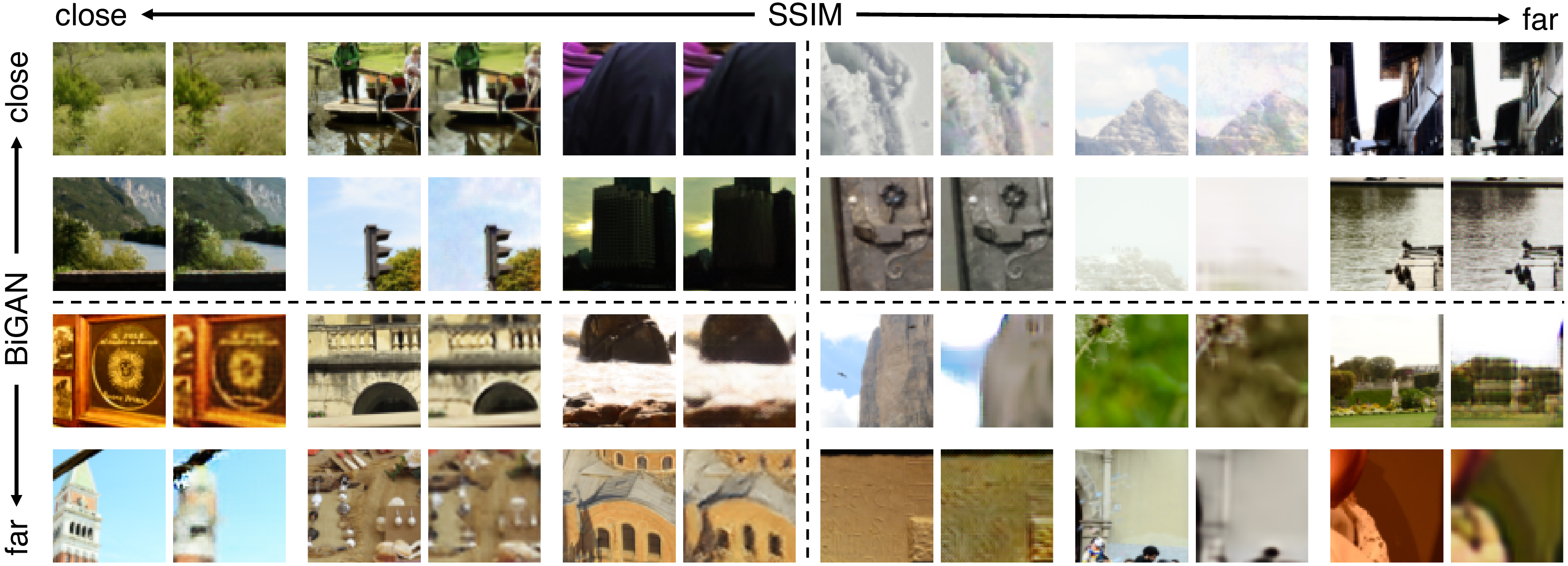}
\end{subfigure}
\vspace{-2mm}
\caption{\textbf{Qualitative comparisons on distortions.} We show qualitative comparison on CNN-based distortions, using the SSIM~\cite{wang2004image} metric and BiGAN network~\cite{donahue2016adversarial}. We show examples where both agree the patches are closer or far, and examples where the metrics disagree. A primary difference is that deep embeddings appear to be more sensitive to blur. Please see the appendix for additional examples.}
\label{fig:qual_comp}
\vspace{-4mm}
\end{figure*}

In Table~\ref{tab:res_quant}, we show full quantitative results across all validation sets and considered metrics, including low-level metrics, along with random, unsupervised, self-supervised, supervised, and perceptually-learned networks.

In Figures~\ref{fig:quant1}, ~\ref{fig:quant2}, ~\ref{fig:quant3}, we plot performance in individual validation sets. Figure~\ref{fig:quant1} shows our traditional and CNN-based distortions, and Figures~\ref{fig:quant2},~\ref{fig:quant3} show results on real algorithm applications individually.

\paragraph{Human performance} If humans chose patches \{$x_1$,$x_0$\} with fraction \{$p$,$1-p$\}, the theoretical maximum for an oracle is $\max(p,1-p)$. However, human performance is lower. If an agent chooses them with probability \{$q$,$1-q$\}, the agent will agree with $qp + (1-q)(1-p)$ humans on expectation. With a human agent, $q=p$, the expected human score is $p^2 + (1-p)^2$.

\paragraph{Linearly calibrating networks} Learning linear weights on top of the \tbfit{Alex} model achieves state-of-the-art results on the real algorithms test set. The \tbfit{linear} models have a learned linear layer on top of each channel, whereas the out-of-the-box versions weight each channel equally. In Figure~\ref{fig:weights_learned}, we show the learned weights for the \tbfit{Alex --frozen} model. The \texttt{conv1-5} layers contain 64, 192, 384, 256, and 256 channels, respectively, for a total of 1152 weights. For each layer, \texttt{conv1-5}, $79.7\%$, $71.4\%$, $56.8\%$, $46.5\%$, $27.7\%$, respectively, of the weights are zero. This means that a majority of the \texttt{conv1} and \texttt{conv2} units are ignored, and almost all of the \texttt{conv5} units are used. Overall, about half of the units are ignored. Taking the cosine distance is equivalent to setting all weights to 1 (Figure~\ref{fig:weights_ones}).

\paragraph{Data quantity for training models on distortions} The performance of the validation set on our distortions ($80.6\%$ and $81.4\%$ for \tbfit{Alex -- tune} and \tbfit{VGG -- tune}, respectively), is almost equal to human performance of $82.6\%$. This indicates that our training set size of 150k patch pairs and 300k judgments is nearly large enough to fully explore the traditional and CNN-based distortions which we defined. However, there is a small gap between the \tbfit{tune} and \tbfit{scratch} models ($0.4\%$ and $0.6\%$ for \tbfit{Alex} and \tbfit{VGG}, respectively).

\section{Model Training Details}
\label{sec:train}

We illustrate the loss function for training the network in Figure~\ref{fig:network} (right) and describe it further in the supplementary material. Given two distances, $(d_0, d_1)$, we train a small network $\mathcal{G}$ on top to map to a score $\hat{h}\in(0,1)$. The architecture uses two 32-channel \texttt{FC-ReLU} layers, followed by a 1-channel \texttt{FC} layer and a sigmoid. Our final loss function is shown in Equation~\ref{eqn:loss}.
\begin{equation}
\begin{split}
    \mathcal{L}(x,x_0,x_1,h)= - h \log \mathcal{G}(d(x,x_0),d(x,x_1)) \\ - (1-h) \log (1-\mathcal{G}(d(x,x_0),d(x,x_1)))
\end{split}
\label{eqn:loss}
\end{equation}

In preliminary experiments, we also tried a ranking loss, which attempts to force a constant margin between  patch pairs $d(x,x_0)$ and $d(x,x_1)$. We found that using a learned network, rather than enforcing the same margin in all cases, worked better.

Here, we provide some additional details on model training for our networks trained on distortions. We train with 5 epochs at initial learning rate $10^{-4}$, 5 epochs with linear decay, and batch size 50. Each training patch pair is judged 2 times, and the judgments are grouped together. If, for example, the two judges are split, then the classification target ($h$ in Figure 3) will be set at 0.5. We enforce non-negative weightings on the linear layer $w$, since larger distances in a certain feature should not result in two patches becoming closer in the distance metric. This is done by projecting the weights into the constraint set at every iteration. In other words, we check for any negative weights, and force them to be 0. The project was implemented using \texttt{PyTorch}~\cite{paszkepytorch}.

\begin{figure}[h!]
\centering 
\includegraphics[width=1.\linewidth]{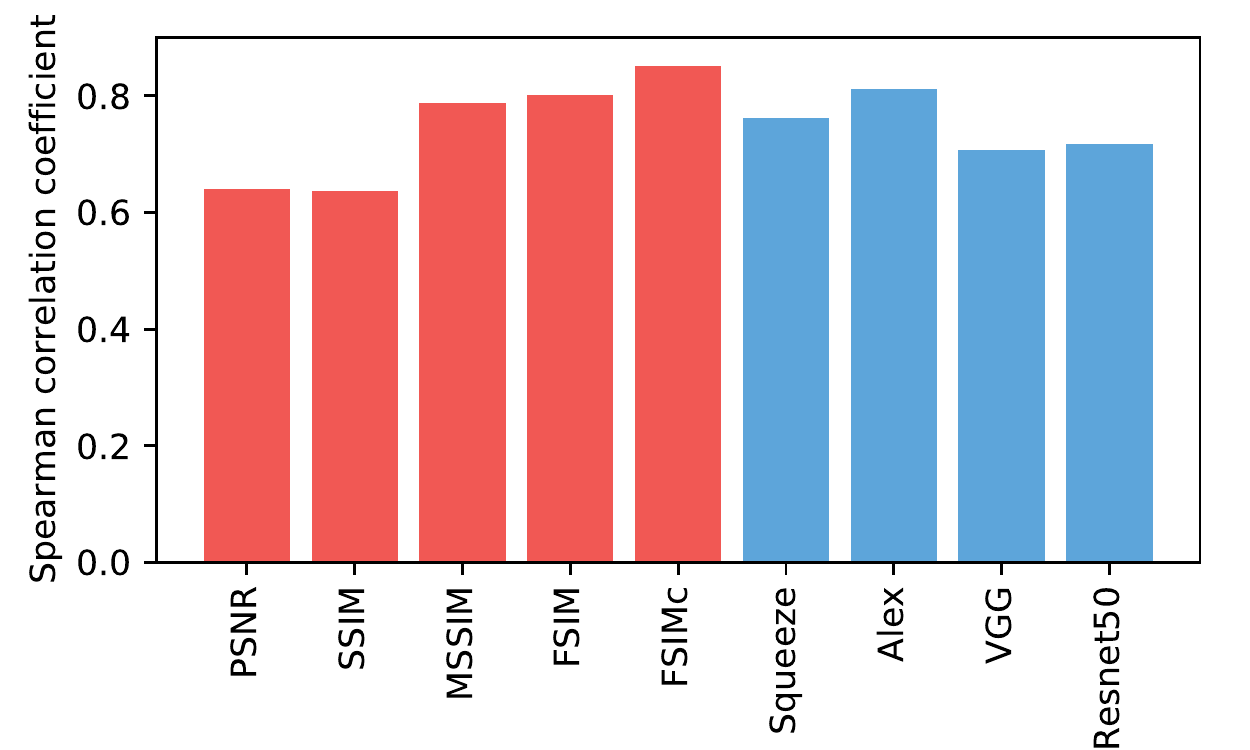} \\
\vspace{-2mm}
\caption{\label{fig:tid}
\textbf{TID Dataset} We show the Spearman correlation coefficient of various methods on the TID2013 Dataset~\cite{ponomarenko2015image}. Note that deep networks trained for classification perform well out of the box (blue).
}
\vspace{-3mm}
\end{figure}

\section{TID2013 Dataset}
\label{sec:tid}

In Figure~\ref{fig:tid}, we compute scores on the TID2013~\cite{ponomarenko2015image} dataset. 
We test the images at a different resolutions, using $\{128, 192, 256, 384, 512\}$ for the smaller dimension. 
We note that even averaging across all scales and layers, with no further calibration, the AlexNet~\cite{krizhevsky2014one} architecture gives scores near the highest metric, FSIMc~\cite{zhang2011fsim}. On our traditional perturbations, the FSIMc metric achieves $61.4\%$, close to $\ell_2$ at $59.9\%$, while the deep classification networks we tested achieved $73.3\%$, $70.6\%$, and $70.1\%$, respectively. The difference is likely due to the inclusion of geometric distortions in our dataset. Despite their frequent use in such situations, metrics such as SSIM were not designed to handle geometric distortions~\cite{sampat2009complex}.

\section{Changelog}
\label{sec:change}
\paragraph{v1} initial preprint release

\paragraph{v2} CVPR camera ready; moved TID results (Appendix~\ref{sec:tid}), SSIM vs BiGAN (Figure~\ref{fig:qual_comp}), and some training details into the Appendix to fit into 8 page limit; clarified that SSIM was not designed to handle geometric distortions~\cite{sampat2009complex} and clarified that our dataset is a perceptual similarity dataset (as opposed to an IQA dataset); added linear weights for \textit{Squeeze-lin} and \textit{VGG-lin} architectures in Figure~\ref{fig:weights}; miscellaneous small edits to text.

\end{document}